\documentclass{article}

\PassOptionsToPackage{numbers, sort&compress}{natbib}

\usepackage[final]{neurips_2016} 

\usepackage[T1]{fontenc}
\usepackage[utf8]{inputenc}
\usepackage{amsmath, amssymb, amsthm}
\usepackage{bm}
\usepackage{graphicx}
\usepackage{booktabs}
\usepackage{comment}
\usepackage{booktabs}
\usepackage{array}
\usepackage{threeparttable}
\usepackage{caption}
\usepackage{mathtools}
\usepackage{tikz}
\usepackage{subcaption}
\usepackage{soul}
\usepackage{algorithm}
\usepackage{algorithmic}
\usepackage{minitoc, tocloft}
\usepackage{appendix}
\usepackage{float}
\usepackage{rotating}
\usepackage{fancyhdr}
\usepackage{enumitem}

\fancypagestyle{plain}{%
  \fancyhf{}%
  \fancyfoot[C]{\confnotice\\[2pt]\thepage}%
}

\usepackage[colorlinks=true, citecolor=blue, linkcolor=blue, urlcolor=red]{hyperref}
\usepackage[capitalize]{cleveref}

\newif\ifshowextra
\showextratrue      

\title{Causal multi-modal AI for personalized chemosensitivity prediction}

\newcommand{\affnum}[1]{\textsuperscript{#1}}

\author{%
  \begin{minipage}{\textwidth}
  \centering
  {\bfseries
  Dhruva Biswas\affnum{1},
  Jeroen Berrevoets\affnum{1},
  Alec McClean\affnum{1},
  Linus Bao\affnum{1},
  Jungkyu Park\affnum{1},
  Ken G.~Zeng\affnum{1},\\
  Joseph Cappadona\affnum{1},
  Cerise Tang\affnum{1},
  Chuwen Liu\affnum{1},
  Bartosz Machura\affnum{1},
  Yin Wu\affnum{2,3},
  Valerie Speirs\affnum{4,5},
  Hatem Soliman\affnum{6},\\
  Rohit Bhargava\affnum{7,8},
  Sheheryar Kabraji\affnum{9},
  Thaer Khoury\affnum{9},
  David Page\affnum{10,11},
  Brian Piening\affnum{10,11},
  Carlo Bifulco\affnum{10,11},
  Claudia Meurs\affnum{12},
  Pieter Westenend\affnum{12},
  Sylvie Chabaud\affnum{13},
  Jerome Lemonnier\affnum{14},
  Paul H.~Cottu\affnum{15},
  Florence Dalenc\affnum{16},
  Fabrice Andre\affnum{17},
  Frederique Madeleine Penault-Llorca\affnum{18},
  Thomas Bachelot\affnum{19},
  Frederick Howard\affnum{20},
  \\Francisco J.~Esteva\affnum{21},
  Kevin Kalinsky\affnum{22},
  Lajos Pusztai\affnum{23},
  Jan Witowski\affnum{1},
  Krzysztof J.~Geras\affnum{1,$\dagger$}
  }\\[1.2em]
  {\normalfont\footnotesize
  \affnum{1}Ataraxis AI, New York, NY, USA.\\
  \affnum{2}Centre for Inflammation Biology and Cancer Immunology, King's College London, London, UK.\\
  \affnum{3}Department of Medical Oncology, Guy's Hospital, London, UK.\\
  \affnum{4}School of Medicine, Medical Sciences and Nutrition, University of Aberdeen, Aberdeen, UK.\\
  \affnum{5}Aberdeen Cancer Centre, University of Aberdeen, Aberdeen, UK.\\
  \affnum{6}Department of Breast Oncology, H. Lee Moffitt Cancer Center and Research Institute, Tampa, FL, USA.\\
  \affnum{7}Department of Pathology, University of Pittsburgh, Pittsburgh, PA, USA.\\
  \affnum{8}Magee-Womens Hospital of UPMC, Pittsburgh, PA, USA.\\
  \affnum{9}Roswell Park Comprehensive Cancer Center, Buffalo, NY, USA.\\
  \affnum{10}Providence Genomics, Portland, OR, USA.\\
  \affnum{11}Earle A. Chiles Research Institute, Providence Cancer Institute, Portland, OR, USA.\\
  \affnum{12}Laboratory of Pathology, Dordrecht, Netherlands.\\
  \affnum{13}Department of Clinical Research and Innovation, Centre Léon Bérard, Lyon, France.\\
  \affnum{14}R\&D Unicancer, Paris, France.\\
  \affnum{15}Medical Oncology, Institut Curie, Université, Paris, France.\\
  \affnum{16}Institut Claudius Regaud, IUCT-Oncopole, Toulouse, France.\\
  \affnum{17}Gustave Roussy and Paris-Saclay University, Villejuif, France.\\
  \affnum{18}Centre Jean Perrin, Université Clermont Auvergne, INSERM U1240, Clermont-Ferrand, France.\\
  \affnum{19}Medical Oncology Department, Centre Léon Bérard, Lyon, France.\\
  \affnum{20}University of Chicago, Chicago, IL, USA.\\
  \affnum{21}Northwell Health, New York, NY, USA.\\
  \affnum{22}Winship Cancer Institute of Emory University, Atlanta, GA, USA.\\
  \affnum{23}Yale Cancer Center, New Haven, CT, USA.\\[0.6em]
  \affnum{$\dagger$}Correspondence to: Krzysztof J.~Geras \texttt{<krzysztof.geras@ataraxis.ai>}.
  }
  \end{minipage}%
}

\makeatletter
\def\@notice{}
\makeatother

\begin{document}
\doparttoc
\faketableofcontents
\newgeometry{
  textheight=9in,
  textwidth=6.9in,
  top=1in,
  headheight=12pt,
  headsep=13pt,
  footskip=25pt
}
\setlength{\headwidth}{\textwidth}   

\maketitle
\begin{abstract}
Chemotherapy improves survival for some patients with breast cancer, but doctors cannot reliably predict who.
Current guidelines rely on recurrence scores as a proxy for treatment benefit, which may contribute to the overprescription of chemotherapy.
Here we present a causal multi-modal AI model that predicts personalized chemosensitivity using routinely collected pathology and clinical information.
We developed our model on a multi-national dataset of 9,141 patients (twelve cohorts, nine countries) and evaluated it on another 1,994 patients (five cohorts, three countries).
The model generated treatment-specific recurrence probabilities for each patient, with near-perfect calibration and strong prognostic discrimination across both 5- and 10-year follow-up horizons.
Moreover, its chemotherapy benefit predictions demonstrated robust predictive performance, and out-performed existing recurrence-score-based tests.
Compared to the standard of care, using the model to support personally tailored therapeutic decisions could reduce the number of patients receiving chemotherapy by 30\% while achieving the same recurrence-free rate.
Tumors predicted to be highly chemosensitive displayed concordant molecular and morphological programs of proliferation, cell cycle progression, and replication stress.
The model's predictive capabilities transferred zero-shot to non-breast cancers, indicating our causal multi-modal AI approach may provide a universal strategy to predict treatment outcomes across cancer types.

\end{abstract}

\newpage

\section{Introduction}

Breast cancer is the most commonly diagnosed cancer in women worldwide~\citep{GBD_TheLancet_2025}. Hormone-receptor--positive, human epidermal growth factor receptor 2--negative (HR+/HER2$-$) breast cancer accounts for approximately 70\% of cases~\citep{Howlader_JNCI_2014}, with adjuvant endocrine therapy (ET) indicated for all patients~\citep{NCCN_BreastCancer_2026}. Landmark trials have demonstrated that the addition of adjuvant chemotherapy --- chemoendocrine therapy (CET) --- further improves survival outcomes~\citep{Albain_Lancet_2009, Fisher_JNCI_1997}. However, the survival benefit of chemotherapy varies widely between individuals~\citep{EBCTCG_Lancet_2005}. Moreover, the benefit of chemotherapy, which may be marginal, must be carefully weighed against treatment-related toxicities~\citep{Curigliano_TheBreast_2016}. Therefore, accurate personalized estimation of adjuvant chemotherapy benefit is critical to guide therapeutic decision-making in patients with early breast cancer. 

Precision oncology aims to address this challenge~\citep{vanTVeer_Nature_2008, Prat_NatRevClinOncol_2012}, thereby reducing overtreatment in breast cancer~\citep{katz_jama_2018}. 
Clinical guidelines recommend the use of genomic assays designed to estimate the risk of recurrence, to inform chemotherapy decision-making~\citep{NCCN_BreastCancer_2026}. 
All commercially available assays were developed and first validated as prognostic tools: in trial cohorts receiving ET alone for the 21-gene recurrence score (Oncotype DX; NSABP B-14~\citep{Paik_NEJM_2004}) and the 50-gene risk-of-recurrence score (PAM50/Prosigna; ATAC and ABCSG-8~\citep{Dowsett_JCO_2013, Gnant_AnnOncol_2014}), and in a consecutive single-institution series with heterogeneous adjuvant treatment for the 70-gene signature (MammaPrint)~\citep{vandeVijver_NEJM_2002}.
Subsequently, retrospective analyses of randomized trials suggested that the 21-gene recurrence score was associated with chemotherapy benefit~\citep{Paik_JCO_2006, Albain_LancetOncol_2010}, and large prospective trials then indicated that chemotherapy de-escalation in patients classified as low genomic risk did not, on average, worsen outcomes: TAILORx and RxPONDER for the 21-gene assay in node-negative and node-positive disease respectively~\citep{Sparano_NEJM_2018, Kalinsky_NEJM_2021}, MINDACT for the 70-gene assay in clinically high-risk patients~\citep{Cardoso_NEJM_2016, Piccart_LancetOncol_2021}, and, most recently, OPTIMA for the 50-gene assay in clinically high-risk patients with up to nine positive nodes~\citep{Stein_JCO_2026}. 
However, recurrence risk and chemosensitivity are two distinct properties of a tumor~\citep{ballman2015biomarker, aldea2025esmo}. 
Consequently, patients with identical genomic risk can exhibit markedly different treatment benefits~\citep{clark2008prognostic, oldenhuis2008prognostic}. 
Moreover, therapeutic decision-making based on chemotherapy de-escalation in patients at low genomic risk cannot distinguish high-risk patients who will benefit from chemotherapy from those who will not --- the latter being candidates for escalation with novel therapeutics, such as CDK4/6 inhibitors~\citep{johnston2020abemaciclib, slamon2024ribociclib}.
Access to commercially available prognostic tests is also limited by the cost and tissue demands of the molecular testing they require~\citep{mateo2022delivering}.

In parallel, advances in artificial intelligence (AI), particularly deep learning, have enabled the creation of models which can extract compact and useful representations from high-dimensional data~\citep{LeCun_nature_2015}. In medicine, these models have been shown to learn and exploit subtle visual patterns beyond the limits of human perception~\citep{biswas_jaccadv_2025, witowski_natcomms_2026, esteva_nature_2017, poterucha_nature_2025, makino2022differences}. In computational pathology they have enabled prediction of disease state from a single routinely collected hematoxylin-and-eosin (H\&E)-stained whole slide image (WSI)~\citep{beck_stm_2011, coudray_naturemed_2018, kather_naturemed_2019, campanella_naturemed_2019, bera_natrevclinoncol_2019, marra_annalsoncology_2025}. Pathology foundation models have advanced these capabilities, leveraging self-supervised learning on massive unlabeled datasets to produce robust feature extractors that generalize across domains without fine-tuning~\citep{cappadona_neuripsworkshop_2024, chen_naturemed_2024, xu_nature_2024, lu_naturebme_2021, wang_nature_2024}. As this pretraining does not require labels, representations can be learned from very large image collections, and models built on them need far fewer labelled examples than training from scratch would require~\citep{lu_naturebme_2021, bommasani2021}, supporting a range of clinical prediction tasks~\citep{lu_nature_2021, vorontsov_natmed_2024}. However, in breast cancer, AI pathology models have remained purely prognostic, directly associating with clinical outcomes~\citep{witowski_natcomms_2026, sharma_breastcancerres_2024, fernandez_breastcancerres_2022, amgad_natmed_2024, garberis_natcommun_2025} or mimicking the result of prognostic gene expression assays~\citep{boehm_naturecomms_2024, shamai_lancetoncology_2025, cohen_npjbreast_2026}. Neither of these approaches yields a direct estimate of chemotherapy benefit, limiting the clinical utility of current AI models for therapeutic decision-making.

\begin{figure}[h]
    \centering
    \vspace{-1cm}
    \includegraphics[width=1.0\linewidth]{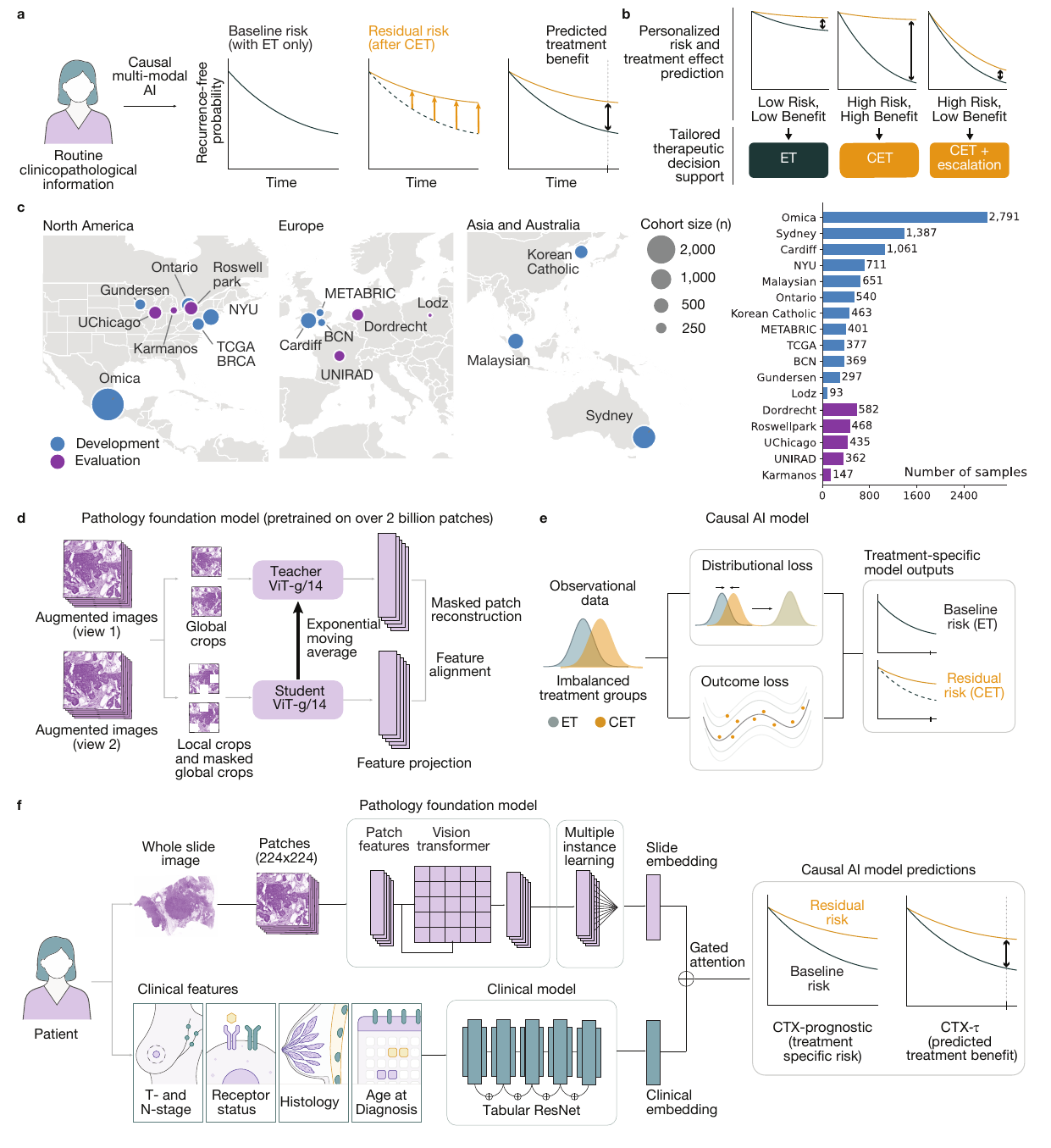}
    \caption{
    \scriptsize
    \textbf{Causal multi-modal AI model for personalized chemosensitivity prediction.} 
    \textbf{a}, Survival curves are shown for a hypothetical patient with HR+/HER2$-$ breast cancer to illustrate baseline risk with endocrine therapy (ET, green line) and residual risk with chemoendocrine therapy (CET, gold line). The difference is the personalized chemotherapy benefit (black arrow).
    \textbf{b}, Three scenarios for how prognosis and chemotherapy benefit predictions could provide personally tailored therapeutic decision support for patients with HR+/HER2$-$ breast cancer: (1) low risk and low-benefit, for whom ET is sufficient; (2) high risk and high-benefit, for whom CET is warranted; (3) high risk but low-benefit from standard chemotherapy, for whom escalation with additional agents may be required.
    \textbf{c}, Development and evaluation cohorts. Marker area is proportional to cohort size. Blue indicates the development set (12 cohorts, n = 9,141) and purple the held-out evaluation set (5 cohorts, n = 1,994).
    \textbf{d}, Pathology foundation model training. The foundation model learns general-purpose representations of tissue morphology through self-supervised pretraining on over two billion H\&E image patches. Training follows a similar procedure to DINOv2~\citep{oquab2024dino}, a state-of-the-art self-supervised learning method in which the model is taught to produce consistent representations of the same tissue region viewed under different augmentations, using a giant vision transformer (ViT-g). Falcon, the pathology foundation model used here, is a successor to Kestrel~\citep{cappadona_neuripsworkshop_2024}.
    \textbf{e}, Causal AI model training. A counterfactual regression network (CFRNet~\citep{shalit_icml_2017, johansson_icml_2016}) uses a maximum mean discrepancy penalty to balance the representation distributions of the ET and CET treatment groups, then predicts recurrence separately under each treatment.
    \textbf{f}, Architecture of the multi-modal AI. Whole-slide images are divided into 224$\times$224 patches, encoded by the pathology foundation model, and aggregated by multiple instance learning into a single slide-level embedding; clinical features are encoded separately by a tabular residual network. The two embeddings are fused with a gated attention layer and passed to CFRNet, the causal model. The causal model outputs CTX-prognostic --- the predicted risk of recurrence under the treatment a patient actually received, baseline risk under ET or residual risk under CET --- and CTX-$\tau$, the predicted increase in recurrence-free probability with CET versus ET, which is the personalized estimate of chemotherapy benefit.
    }
    \label{fig:figure_1}
\end{figure}

Causal AI models go beyond prognosis by predicting treatment-specific recurrence probabilities~\citep{feuerriegel_natmed_2024, kunzel_pnas_2019}, for example under ET versus CET in HR+/HER2$-$ breast cancer (\textbf{\Cref{fig:figure_1}a}). The difference between these two predictions gives an estimate of personalized chemotherapy benefit, which can be used for tailored therapeutic decision support (\textbf{\Cref{fig:figure_1}b}). AI models built for this task learn balanced representations across treatment groups, essentially making the covariates in each treatment group comparable, such that a treatment-specific prediction is both accurate and unbiased~\citep{kyriacou_jama_2016, shalit_icml_2017, johansson_icml_2016}. Applied to routine clinical variables together with morphological features extracted from histopathology images by a pathology foundation model, we reasoned that this approach may yield a predictive biomarker from multi-modal data.

In this paper, we introduce a causal multi-modal AI model (CTX) that predicts personalized chemosensitivity in HR+/HER2$-$ early breast cancer. We assemble a multi-national observational dataset of 11,135 patients spanning ten countries (\textbf{\Cref{fig:figure_1}c}). Next, we develop CTX by integrating clinical characteristics with foundation-model-derived pathology representations, and learn chemotherapy benefit with causal adjustment (\textbf{\Cref{fig:figure_1}d--f}). We then rigorously evaluate CTX across five axes: prognostic accuracy, predictive performance, clinical utility, explainability, and generalizability. 

\clearpage

\section{Results}

\subsection{Multi-national development and evaluation datasets} \label{subsec:data-description}

We assembled a multi-national observational dataset of 11,135 patients with early breast cancer sourced from 17 cohorts (\textbf{\Cref{fig:figure_1}c}, \textbf{Extended Data \Cref{tab:population}}). We divided this dataset into a development set derived from twelve cohorts (n = 9,141) and a held-out evaluation set of HR+/HER2$-$ patients from five cohorts (n = 1,994, \textbf{Extended Data \Cref{tab:test}}). Clinical covariates and at least one de-identified WSI were collected for each patient. Clinicopathological features were harmonized across all cohorts. Treatment information, including chemotherapy status, was curated for all patients; 47.4\% received CET in the development set, and 40.7\% in the evaluation set. The primary clinical endpoint was recurrence-free interval (RFI), defined as the time from diagnosis to the first local-regional or distant recurrence, with death treated as a censoring event~\citep{hudis_jco_2007, tolaney_jco_2021}.

Because chemotherapy assignment in these observational cohorts was not randomized, patients receiving CET versus ET differ systematically in prognostic factors. Therefore, a naive CET-versus-ET comparison could misattribute variation in baseline risk for treatment benefit~\citep{kyriacou_jama_2016}. To address this, we estimated the propensity scores for patients in the evaluation set using their clinical covariates, and reweighted with inverse propensity weights (IPW) throughout the evaluation analyses~\citep{cole2004adjusted, austin2011introduction}. IPW-adjustment reduced the absolute average standardized mean difference (SMD) between CET and ET patients in the evaluation set from 0.376 to 0.125 (\textbf{Extended Data \Cref{fig:propensity_scores}}), below the 0.20 threshold conventionally regarded as negligible imbalance~\citep{cohen2013statistical}. This improved covariate balance ensures that the predictive evaluation accounts for baseline prognostic differences between treatment groups.

\subsection{CTX: a causal multi-modal AI model for estimating personalized chemotherapy benefit} \label{subsec:ctx}

Here, we briefly describe the construction of CTX, a causal multi-modal AI model developed to predict recurrence risk and chemotherapy benefit in HR+/HER2$-$ early breast cancer (\textbf{\Cref{fig:figure_1}d--f}). Further details are presented in the Methods (\textbf{\cref{sec:methods}}) and Supplementary Methods (\textbf{\cref{sec:methods:ctx}}).

CTX was built from two components: a multi-modal network, which integrates histopathology slides and clinical features into a single latent representation, and a counterfactual regression network (CFRNet~\citep{shalit_icml_2017, johansson_icml_2016}).
The multi-modal network combines WSIs --- embedded using Falcon, our pathology foundation model and successor to Kestrel~\citep{cappadona_neuripsworkshop_2024} --- and clinical features.
Falcon was pretrained via self-supervised learning on over two billion WSI patches, producing a strong, general purpose patch-level encoder (\textbf{\Cref{fig:figure_1}d}). 
Falcon has demonstrated state-of-the-art performance on a large suite of patch- and slide-level tasks, validating the high quality of the representations it produces (\textbf{Extended Data~\Cref{fig:falcon}}).
During the development of CTX, Falcon was frozen and used to extract representations for all patches for a given input slide, which were subsequently aggregated into a single pathology representation via multiple instance learning.
Clinical features were encoded in parallel using a tabular residual network (ResNet~\citep{gorishniy2021revisiting}), producing a single clinical representation.
To yield a multi-modal representation, the pathology and clinical representations were fused using a gated attention module, which was then input to CFRNet. 
Therefore, each CTX prediction draws on pathological information and clinical features, both collected routinely in clinical practice.

CFRNet addresses a challenge that existing prognostic assays were not designed to: estimating the difference in recurrence risk under two treatments. 
Chemotherapy assignment in the development set is not random, so patients receiving ET and CET differ systematically in their clinical features, pathological information, and baseline risk.
As a result, a naive model trained on this data can predict recurrence accurately under the treatment a patient received by relying on features that merely separate ET from CET patients, but may do poorly when predicting recurrence for the alternative ("counterfactual") treatment. 
CFRNet instead learns a representation in which the ET and CET groups resemble one another, so that recurrence can be predicted from features that carry over to both treatments --- keeping each treatment-specific prediction accurate for every patient (\textbf{\Cref{fig:figure_1}e}). 
As a crucial element of our strategy, CFRNet learned its representation directly from the fused multi-modal representation.
On top of this representation, two survival models were fit to predict five-year RFI under each treatment scenario (\textbf{\Cref{fig:figure_1}f}).
This results in two predictions for every patient, under CET versus ET.

These define CTX's two outputs. CTX-prognostic is CTX's predicted risk of recurrence under the treatment a patient actually received, baseline risk under ET or residual risk under CET. 
CTX-$\tau$ is the predicted absolute increase in recurrence-free probability with CET versus ET, that is, the personalized estimate of chemotherapy benefit.
We stratified patients into CTX-$\tau$ high- and low-benefit categories using a 2\% predicted chemotherapy benefit threshold ($\geq$ 2\% = high-benefit; see Methods).
This value lies between the 1\% benefit patients report as sufficient to make chemotherapy worthwhile~\citep{duric2005patients} and the 3--5\% an oncologist consensus panel judged necessary to recommend it~\citep{burstein2025tailoring}.
Notably, this threshold serves only to define benefit categories for reporting; the continuous score allows a clinician to apply whichever benefit threshold they and the patient prefer.

\begin{figure}[h]
    \centering
    \includegraphics[width=1.0\linewidth]{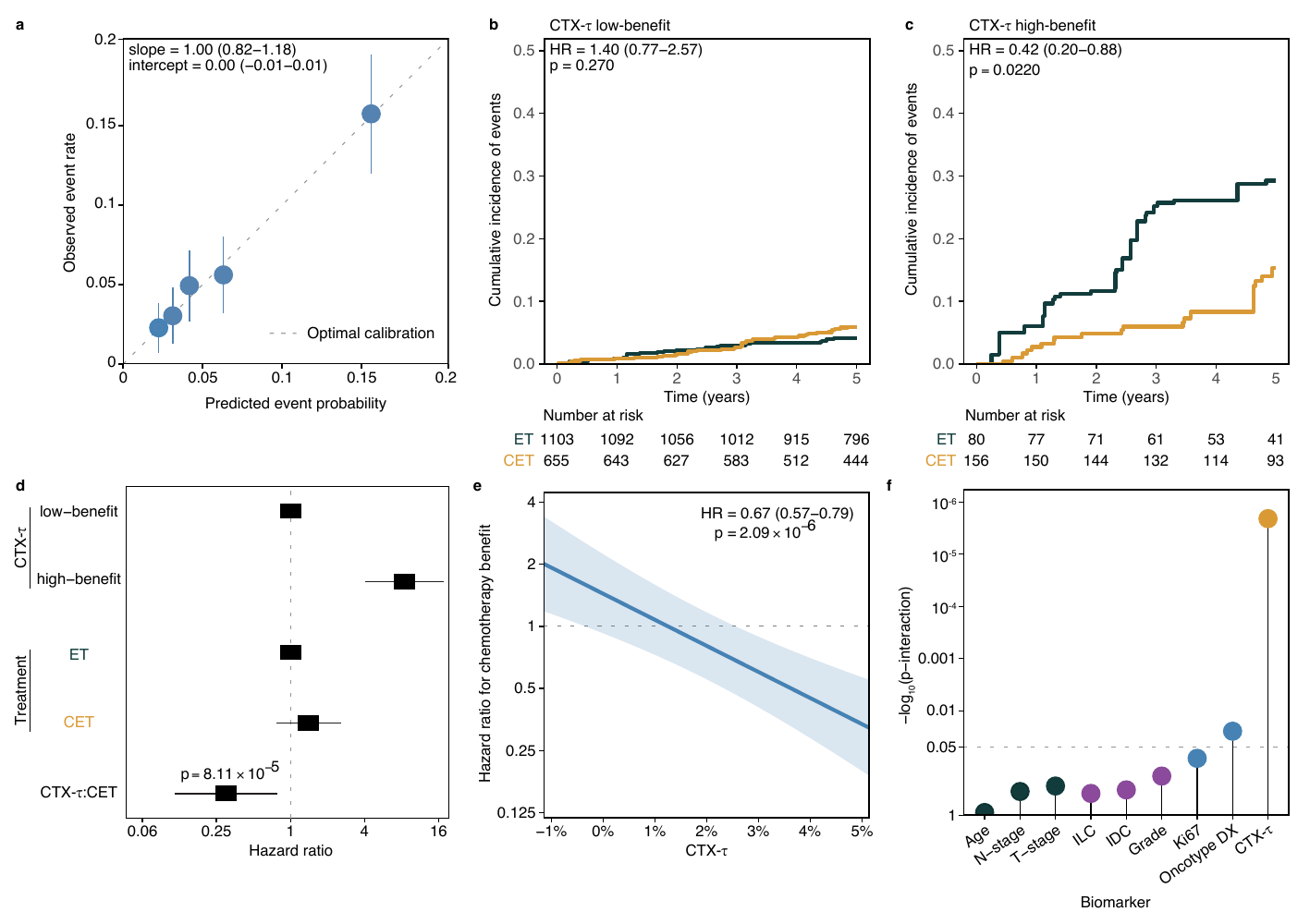}
    \scriptsize
    \caption{
        \textbf{Evaluation of CTX as a predictor of prognosis and chemosensitivity.}
        Evaluations performed in the evaluation set (n=1,994). 
        \textbf{a}, Calibration plot compares CTX-prognostic predictions versus observed event probability for five-year recurrence-free interval (RFI). Circles denote observed event probabilities within binned strata of predicted risk, with vertical error bars representing 95\% CIs; the diagonal indicates perfect calibration.
        \textbf{b, c}, Kaplan--Meier curves for the probability of meeting the RFI endpoint in the CTX-$\tau$ predicted low-benefit (\textbf{b}) and high-benefit (\textbf{c}) groups, stratified by adjuvant therapy (chemoendocrine therapy [CET, gold] vs. endocrine therapy [ET, green]), with risk tables underneath. Risk tables report IPW-adjusted numbers at risk, not raw subgroup sizes. Within each subgroup, the HR, 95\% CI, and p-value are from an IPW-adjusted Cox model containing chemotherapy treatment status as the only covariate, fit to that subgroup; the p-value is from the score test.
        \textbf{d}, Forest plot of the hazard ratio (HR) for CTX-$\tau$, chemotherapy treatment status (ET or CET), and the interaction of chemotherapy and CTX-$\tau$ in an IPW-adjusted Cox model. The center box indicates the HR, and error bars represent 95\% confidence intervals (CIs). p-values are indicated for the treatment-by-biomarker interaction term.
        \textbf{e}, HR for chemotherapy benefit across the range of continuous CTX-$\tau$, with 95\% CI (shaded region). Estimates were derived by the delta method applied to an IPW-adjusted Cox model including an interaction term between continuous CTX-$\tau$ score and chemotherapy treatment, with robust standard errors. The dashed reference line indicates no chemotherapy benefit (HR = 1).
        \textbf{f}, Lollipop plot comparing the predictive performance of clinical, pathological, molecular, and CTX-$\tau$ for five-year RFI. For each marker, the vertical axis shows the treatment-by-biomarker interaction p-value from a Cox model containing the biomarker, chemotherapy treatment status, and their interaction. The dashed horizontal line indicates p = 0.05. Points are colored by feature category: patient (green), clinical (purple), molecular (blue), and CTX (gold).
    }
    \label{fig:validation_predictive}
\end{figure}

\subsection{Evaluation of CTX for predicting prognosis} \label{sec:results:prognostic}

We evaluated the model's ability to predict recurrence in the held-out five-cohort evaluation set (n = 1,994) by comparing the predictions of CTX-prognostic against observed RFI outcomes. 
The ability of the model to rank patients by risk of recurrence was assessed using five-year RFI, the endpoint the model was trained with in the development set. CTX-prognostic demonstrated good discrimination, with a time-dependent area under the receiver operating characteristic curve (AUC) of 0.711 (95\% CI = [0.663, 0.756], \textbf{Extended Data  \Cref{fig:validation_prognostic}a}).
It further generalized to 10-year RFI (AUC = 0.765, 95\% CI = [0.708, 0.791]), suggesting that the model captures temporally consistent prognostic signals. 
To determine whether CTX-prognostic provided additional information beyond clinical features known to predict recurrence risk, we conducted a multivariable Cox analysis adjusting for age, tumor stage, and nodal stage. CTX-prognostic remained an independent predictor of RFI at both 5 years (adjusted hazard ratio [HR] = 2.02, 95\% CI = [1.44, 2.83], p = 4.65$\times \text{10}^{-\text{5}}$) and 10 years (adjusted HR = 2.28, 95\% CI = [1.73, 3.00], p = 5.74$\times \text{10}^{-\text{9}}$) (\textbf{Extended Data \Cref{fig:validation_prognostic}b}).
Lastly, we wished to assess model calibration --- the extent to which predicted event probabilities matched the observed event rates. CTX-prognostic's five-year recurrence probability predictions exhibited near-perfect calibration across the full range of predicted risk (slope = 1.00, 95\% CI = [0.82, 1.18]; intercept = 0.00, 95\% CI = [$-$0.01, 0.01]; \textbf{\Cref{fig:validation_predictive}a}).

Together, these results demonstrate that CTX-prognostic performed well as a recurrence predictor, generalizing robustly to a multi-national evaluation set comprising five independent cohorts, across both 5- and 10-year follow-up horizons.

\subsection{Evaluation of CTX for predicting chemotherapy benefit} \label{sec:results:predictive}

We evaluated CTX-$\tau$ as a predictive biomarker of chemosensitivity in the held-out five-cohort evaluation set (n = 1,994) using five-year RFI, the endpoint used for model training in the development set. 
We first evaluated CTX-$\tau$ predictions dichotomized into high- and low-benefit categories. High-benefit patients are those for whom CTX-$\tau$ recommends CET, while low-benefit patients are those for whom CTX-$\tau$ recommends ET. 
Among CTX-$\tau$ low-benefit patients, recurrence rates did not significantly differ between patients receiving CET and ET (HR = 1.40, 95\% CI = [0.77, 2.57], p = 0.270, \textbf{\Cref{fig:validation_predictive}b}). By contrast, in the CTX-$\tau$ high-benefit category, patients receiving CET exhibited significantly lower recurrence rates than those receiving ET (HR = 0.42, 95\% CI = [0.20, 0.88], p = 0.022, \textbf{\Cref{fig:validation_predictive}c}). 

To further evaluate the ability of CTX-$\tau$ to distinguish between patients who benefited from chemotherapy and those who did not, we examined treatment-by-biomarker interactions in a Cox model --- the gold standard test for assessing predictive performance~\citep{ballman2015biomarker} --- accounting for confounding in observational data via IPW-adjustment~\citep{austin2011introduction}.
The treatment-by-biomarker interaction for CTX-$\tau$ was significant (p = 8.11$\times \text{10}^{-\text{5}}$, \textbf{\Cref{fig:validation_predictive}d}), indicating strong utility for predicting chemotherapy benefit. 
This interaction remained significant in a Cox model after adjustment for age, tumor stage, nodal stage, grade, and histology (IDC/ILC) without propensity weighting (p = 2.39 $\times \text{10}^{-\text{3}}$, \textbf{Extended Data \Cref{fig:supp_benefit}a}). 

To test whether predictive performance was an artifact of the 2\% threshold used to define CTX-$\tau$ categories, we modeled CTX-$\tau$ as a continuous covariate in an IPW-adjusted Cox model. The hazard ratio for chemotherapy benefit decreased monotonically with CTX-$\tau$ score, with a significant treatment-by-biomarker interaction (HR per standard deviation = 0.67, 95\% CI = [0.57, 0.79], p = 2.09 $\times \text{10}^{-\text{6}}$, \textbf{\Cref{fig:validation_predictive}e}), indicating that benefit scaled continuously with the biomarker rather than depending on any specific threshold.

We performed an ablation analysis to quantify the contribution of each input modality to the predictive signal (\textbf{Extended Data \Cref{fig:supp_benefit}b}). 
The treatment-by-biomarker interaction was significant for both the clinical-only score (p = 4.41 $\times \text{10}^{-\text{4}}$) and the pathology-only score (p = 7.90 $\times \text{10}^{-\text{3}}$). 
Importantly, adding the pathology-derived interaction to a model already containing the clinical-derived interaction significantly improved fit (likelihood ratio test p = 9.79 $\times \text{10}^{-\text{3}}$), indicating that histopathology captures chemosensitivity information that is not recoverable from clinical features alone.

Lastly, to contextualize the predictive value of CTX-$\tau$, we benchmarked it against a panel of established clinical, pathological, and molecular biomarkers (\textbf{\Cref{fig:validation_predictive}f}). 
Each biomarker was standardized to zero mean and unit standard deviation. 
We then fitted an IPW-adjusted Cox model with treatment, the standardized biomarker, and their interaction using all available data.
The treatment-by-biomarker interaction thus reflected the change in chemotherapy benefit per standard deviation of each biomarker on a common scale. 
Most clinicopathological features (age, tumor stage, nodal stage, tumor grade, histological subtype [IDC/ILC], and Ki67 status) showed no significant treatment-by-biomarker interaction.
The Oncotype DX recurrence score showed a significant treatment interaction (p = 0.0247), consistent with its known association with chemotherapy sensitivity.
Across the full panel, CTX-$\tau$ displayed the most statistically robust treatment-by-biomarker interaction (p = 2.09 $\times \text{10}^{-\text{6}}$). 
As Oncotype DX scores were only available in a subset of the evaluation cohort (n = 983), we performed a strict head-to-head comparison within the same patients: CTX-$\tau$ retained a stronger treatment-by-biomarker interaction (p = 9.68 $\times \text{10}^{-\text{4}}$) than the Oncotype DX recurrence score (p = 0.0247, \textbf{Extended Data \Cref{fig:supp_benefit}c}).

Overall, CTX-$\tau$ exhibited a significant treatment-by-biomarker interaction --- stratifying patients into categories with markedly different chemotherapy benefit --- and showed the strongest interaction in a head-to-head comparison against approved chemotherapy-decision biomarkers. These results indicate that the personalized chemosensitivity predictions yielded by CTX-$\tau$ may improve therapeutic decision-making.

\subsection{Personally tailored therapeutic decision support} \label{sec:results:decision}

Unlike existing risk-based assays, which conflate prognosis with chemosensitivity, CTX separates these quantities into CTX-prognostic and CTX-$\tau$. Here, we explored whether CTX-$\tau$ uncovered heterogeneity missed by CTX-prognostic and other risk scores, and the potential implications for therapeutic decision-making.

\begin{figure}[h]
    \centering
    \includegraphics[width=1\linewidth,trim={0cm 0 0cm 0},clip]{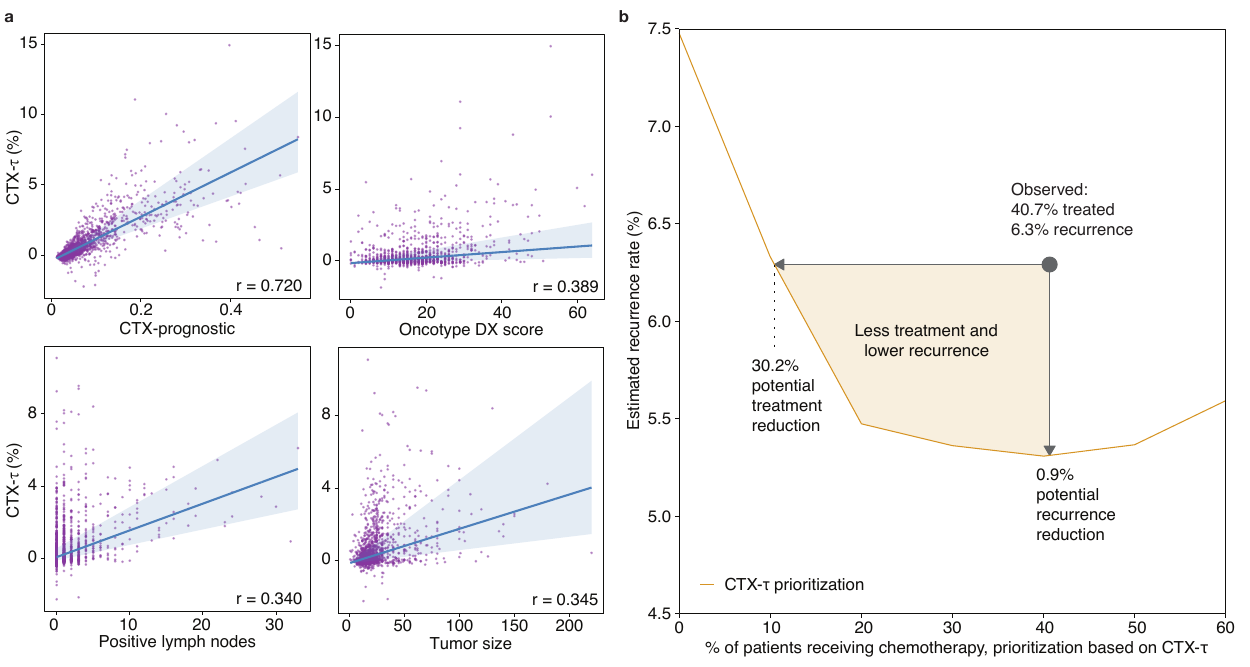}
    \caption{\textbf{CTX separates chemosensitivity from risk to enable personalized therapeutic decision support.}
    \textbf{a}, Scatter plots compare CTX-$\tau$ against four risk estimates: CTX-prognostic, tumor size, number of positive lymph nodes, and Oncotype DX recurrence score. The blue line indicates the median (50th-percentile) quantile regression fit, with the shaded ribbon representing the interquartile range (25th--75th percentile) band. Pearson correlation coefficients (r) are shown for each comparison.
    \textbf{b}, Predicted five-year recurrence rate as a function of the proportion of patients receiving chemotherapy (CET rather than ET). Patients are ranked according to those most likely to benefit from chemotherapy, as judged by CTX-$\tau$ predictions. The curve shows the IPW-adjusted Kaplan--Meier estimated outcomes if the proportion of patients receiving chemotherapy increases from 0\% to 60\%, starting with the patient with the greatest predicted benefit. The x-axis ends at 60\%, within one standard deviation of the mean observed treatment rate across cohorts. The proportion treated with CET and recurrence rate is indicated for the observed population (grey circle). The arrows indicate the same values for CTX-$\tau$-guided strategies reducing chemotherapy allocation while matching the recurrence-free rate (horizontal arrow), or improving recurrence-free rate while matching the observed treatment rate (vertical arrow). 
    }
    \label{fig:individualized}
\end{figure}

We first examined the relationship between CTX-$\tau$ score and prognostic estimates in the evaluation set (n=1,994). We compared CTX-$\tau$ to CTX-prognostic, tumor size, number of positive lymph nodes, and Oncotype DX recurrence score (\textbf{\Cref{fig:individualized}a}). CTX-$\tau$ was positively correlated with all four, but the correlation was weak for tumor size (Pearson's correlation coefficient [r] = 0.345), number of positive lymph nodes (r = 0.340), and Oncotype DX recurrence score (r = 0.389) --- indicating that none of these prognostic estimates reliably captured the chemosensitivity described by CTX-$\tau$. While a stronger correlation was observed with CTX-prognostic (r = 0.720), since both predictions were made by the same model it is notable that CTX-prognostic failed to capture much of the variation in CTX-$\tau$. 

We next wished to quantify the potential impact of CTX-guided chemotherapy allocation on population outcomes. To do this we ranked patients by CTX-$\tau$ and estimated the recurrence-free rate achieved by assigning CET to the highest-benefit proportion of patients (\textbf{\Cref{fig:individualized}b}). Using this framework, the impact of CTX-guided chemotherapy assignment can be compared to the actual chemotherapy assignment observed in this cohort. The difference can be expressed as the reduction in the proportion of patients receiving chemotherapy to achieve the same recurrence-free rate. In the evaluation set, 40.7\% of patients received chemotherapy, achieving a five-year recurrence-free rate of 93.7\%. The CTX-guided strategy matched that recurrence-free rate while treating only 10.5\% of patients --- a relative reduction in chemotherapy use of 74.2\%. Alternatively, the difference in strategies can be expressed as the improvement in recurrence-free rate expected for a fixed proportion of patients assigned chemotherapy. When treating the same proportion of patients as observed in this cohort, a CTX-guided strategy achieved a reduction in recurrence rate from 6.3\% to 5.4\%, a relative reduction of 14.3\%.

\begin{figure}[!t]
    \centering
    \vspace{-1cm}
    \includegraphics[width=1.0\linewidth]{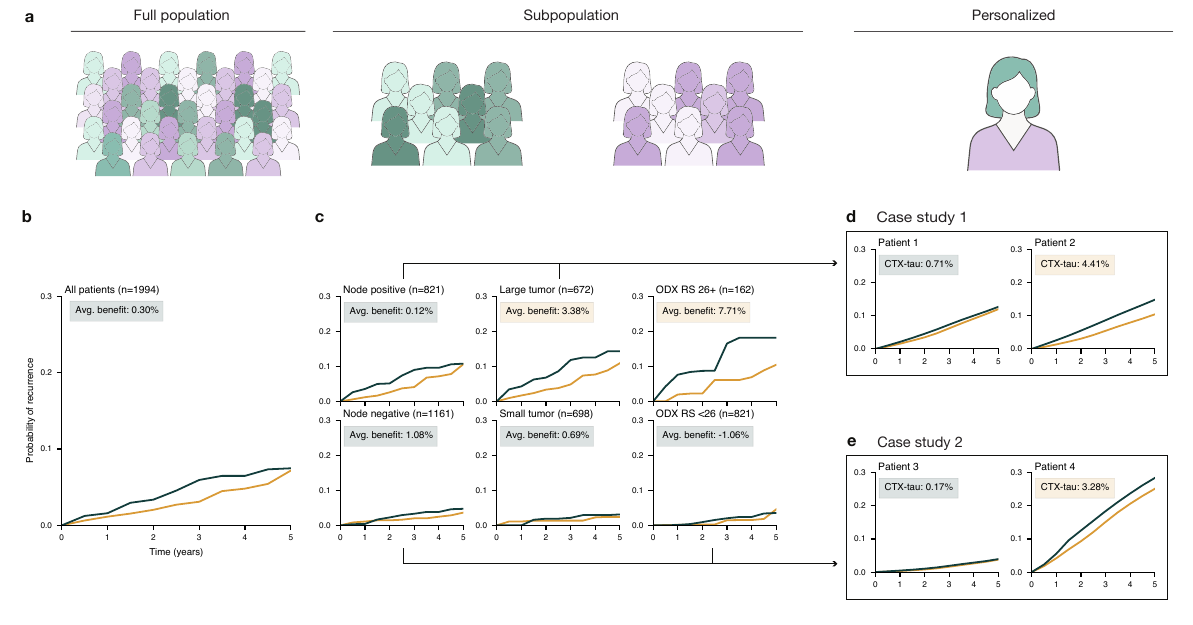}
    \caption{
    \scriptsize
    \textbf{Personalized predictions capture heterogeneity obscured at the (sub)population level.}
    \textbf{a}, Illustration of the hierarchy of populations and subpopulations from which we may estimate prognosis and chemotherapy benefit. Estimates can be derived from the full population (left), subpopulations defined by clinicopathological features (middle), or the individual patient (right). Moving from left to right narrows the group being characterized, recovering precision that coarser estimates average away.
    \textbf{b}, Full population IPW-adjusted Kaplan--Meier (KM) curves estimating the probability of recurrence in the evaluation set (n = 1,994), stratified by adjuvant therapy status (chemoendocrine therapy [CET] in gold vs. endocrine therapy [ET] in green). Average benefit is the difference in the KM-estimated five-year recurrence probability between the two arms (ET minus CET), with positive values denoting lower recurrence with the addition of chemotherapy. The average benefit across the entire evaluation set is 0.30\%.
    \textbf{c}, Subpopulations defined by clinicopathological features: node-positive disease, large tumor (tumor size $\geq$ median), and Oncotype DX category (high-risk RS $\geq$ 26, low- and intermediate-risk RS < 26, the split used for chemotherapy allocation).
    \textbf{d}, Case study 1. 
    Patient 1 and 2 are both premenopausal, node-positive with large tumors.
    The CTX-predicted probability of recurrence under ET and under CET are annotated.
    CTX recurrence probabilities assigned both patients a high baseline risk under ET, in agreement with standard clinicopathological risk indicators. 
    However, CTX-$\tau$ separated them, predicting 0.71\% benefit for patient 1 and 4.41\% benefit for patient 2. 
    In practice, both patients received CET, but their outcomes differed. 
    Patient 1 recurred at approximately 3 years, whereas Patient 2 remained recurrence-free (censored at approximately 3.5 years).
    These patient outcomes are aligned with their CTX-$\tau$ scores, which predicted patient 1 to be significantly less chemosensitive than patient 2.
    \textbf{e}, Case study 2. 
    Patient 3 and 4 both have node-negative disease and intermediate Oncotype DX recurrence scores.
    CTX predicted patient 3 to be at low risk with negligible chemotherapy benefit (CTX-$\tau$ = 0.17\%) and patient 4 to be at high risk with substantial benefit (CTX-$\tau$ = 3.28\%).
    The treatment each patient received ran counter to these predictions: patient 3 received CET and patient 4 received ET alone. 
    Patient 3 remained recurrence-free across approximately 17 years of follow-up, suggesting that chemotherapy de-escalation might have been safe --- as predicted by CTX. Patient 4 recurred at approximately 4.8 years, suggesting that chemotherapy escalation may have been prudent --- as predicted by CTX.
    In both case studies, patients who are indistinguishable based on the clinicopathological features experienced divergent outcomes which were tracked by divergent CTX predictions. 
    Panels \textbf{b}--\textbf{e} share a common y-axis (probability of recurrence) and x-axis (time since diagnosis, years). Individual counterfactual outcomes are never observed, so these examples illustrate the interpretation of CTX-$\tau$ at the level of the individual patient rather than constituting a formal evaluation of predictive accuracy; the latter is presented in \textbf{\Cref{fig:validation_predictive}}.
    }
    \label{fig:personal_km}
\end{figure}

Prognostic risk and treatment benefit are conventionally estimated for populations or coarse subgroups, yet both vary continuously across individuals, and grouping patients into categories discards biologically meaningful variation that population averages cannot recover~\citep{gerstung_naturegenetics_2017, grinfeld2018classification}. Therefore, we examined how estimates of chemotherapy benefit change as the group being characterized narrows from the full population, to clinicopathological subpopulations, to the individual patient (\textbf{\Cref{fig:personal_km}a}).

Across the full evaluation set, the average benefit in five-year recurrence probability from adding chemotherapy was 0.30\% (\textbf{\Cref{fig:personal_km}b}), a population-averaged effect that conceals substantial heterogeneity. Stratifying by clinicopathological features partially resolved this: benefit was larger in large tumors (3.38\%), and high-risk Oncotype DX tumors (RS $\geq$ 26; 7.71\%), and negligible or negative in their counterparts, including low- and intermediate-risk Oncotype DX tumors (RS < 26; $-$1.06\%) (\textbf{\Cref{fig:personal_km}c}). Yet these subgroup averages remained coarse, obscuring divergence between patients who share the same clinicopathological profile. 

We provide two case studies to illustrate how CTX resolved this variation at the level of the individual. In the first case study (\textbf{\Cref{fig:personal_km}d}), two premenopausal, node-positive patients with large tumors were assigned similarly high baseline risk under ET, in agreement with standard risk indicators, but were separated by CTX-$\tau$ (0.71\% vs. 4.41\%); both received CET, and their observed outcomes tracked these predictions, with Patient 1 recurring at approximately 3 years and Patient 2 remaining recurrence-free. In the second case study (\textbf{\Cref{fig:personal_km}e}), two node-negative patients with intermediate Oncotype DX scores --- indistinguishable under current genomic risk stratification --- were separated by CTX into low benefit (Patient 3; CTX-$\tau$ = 0.17\%) and substantial benefit (Patient 4; CTX-$\tau$ = 3.28\%); the treatment each received ran counter to these predictions, and again the observed outcomes aligned with CTX rather than with the shared Oncotype DX score. These examples illustrate that clinicopathological and genomic categories average over patients whose true benefit diverges, and that personalized estimation recovers this variation --- assigning distinct benefit predictions to patients who are otherwise indistinguishable, and doing so in a direction consistent with their observed outcomes. Because individual counterfactual outcomes are never observed, these case studies illustrate the interpretation of CTX-$\tau$ rather than establishing predictive accuracy, which we formally evaluated in \textbf{\Cref{fig:validation_predictive}}.

Together, these results indicate that risk and chemotherapy benefit are distinct axes, that model-guided chemotherapy assignment could substantially reduce overtreatment without compromising outcomes, and that personalized predictions capture heterogeneity obscured at the population level.
Accurate and granular estimates of treatment benefit from a causal multi-modal AI model could enable personally tailored therapeutic decision support.

\subsection{Mechanistic insights into the determinants of chemosensitivity} \label{sec:results:mechanistic}

Doctors are reluctant to act on predictions they cannot explain.
``Black box'' AI models are met with distrust and poor adoption due to lack of insight into what is triggering the model's predictions, even when they outperform standard tools~\citep{lekadir_bmj_2025, parikh_jco_2025}.
To illuminate the biology that may be contributing to CTX predictions, we linked model predictions to morphological and molecular features.

\begin{figure}[h]
    \centering
    \includegraphics[width=1.0\linewidth]{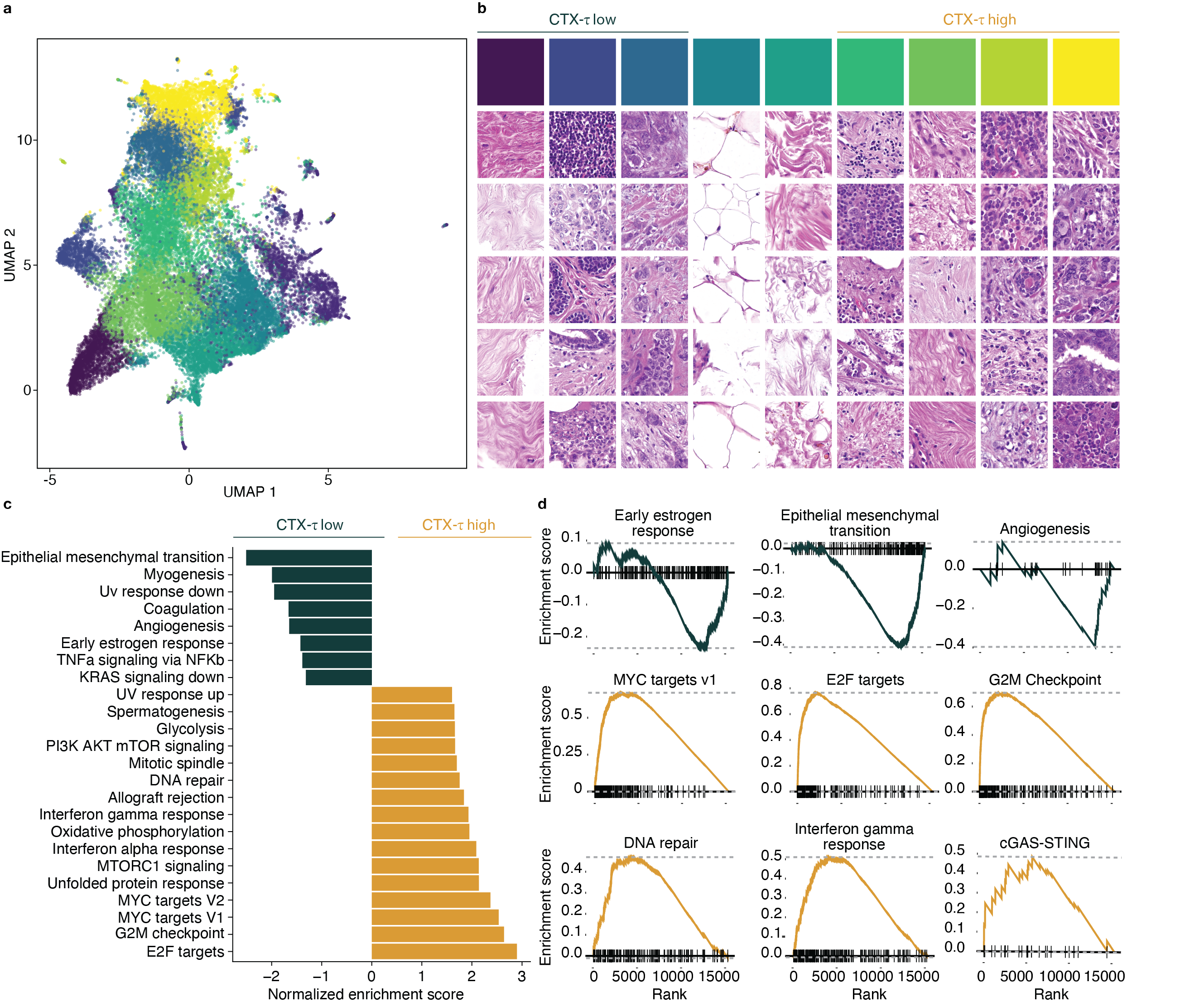}
    \caption{
    \textbf{Morphological and molecular correlates of CTX-$\tau$ predictions.}
    \textbf{a}, Two-dimensional UMAP projection~\citep{mcinnes_arxiv_2018} of embeddings from the pathology foundation model, Falcon, for a cohort in the evaluation set (UNIRAD, n = 362). Each point represents a single H\&E image patch. Patches were partitioned into 9 clusters by $k$-means applied in the embedding space. Clusters are colored by their average CTX-$\tau$ score, from low- (dark blue) to high- (bright yellow) predicted chemotherapy benefit. 
    \textbf{b}, Representative H\&E patches selected from clusters in \textbf{a}, ordered by average CTX-$\tau$ score (low to high).
    \textbf{c}, Bar plot shows gene-set enrichment analysis~\citep{subramanian_pnas_2005} of MSigDB Hallmark pathways for CTX-$\tau$ low- versus high-benefit tumors in the TCGA breast cancer cohort (n=348). Bars show normalized enrichment scores. Only pathways significant at false discovery rate (FDR) < 0.05 are shown.
    \textbf{d}, Pathways enriched in CTX-$\tau$ low-benefit tumors include early estrogen response, epithelial-mesenchymal transition, and angiogenesis. Pathways enriched in CTX-$\tau$ high-benefit tumors include proliferative (E2F targets, G2M checkpoint, MYC targets, MTORC1 signaling), DNA damage response (DNA repair), immune activation (IFN-$\alpha$ and IFN-$\gamma$ response), and cGAS-STING signatures. 
    }
    \label{fig:explainability}
\end{figure}

We first asked what morphological patterns the model associated with chemosensitivity (\textbf{\Cref{fig:explainability}a--b}).
Pathology explainability analysis was conducted within a single representative evaluation cohort (UNIRAD, n = 362).
We segmented the histopathology slide for each patient into patches, then viewed these through the lens of the pathology foundation model by generating embeddings for each patch.
We then grouped the patch embeddings into clusters of morphologically similar patches using $k$-means, and ranked the clusters by the strength of their association with CTX-$\tau$ score.
Three board-certified pathologists with subspecialty expertise in breast pathology, blinded to CTX predictions and clinical outcomes, independently reviewed representative patches from each cluster and annotated the dominant morphological patterns (detailed patch-level annotations are provided in \textbf{Extended Data \Cref{fig:path_explain}}). 
Patches from patients classified as CTX-$\tau$ high-benefit comprised nests of invasive carcinoma, with intermediate-to-high nuclear grade, high cellularity, and identifiable mitotic figures.
In contrast, tumors classified as CTX-$\tau$ low-benefit were dominated by fibrotic stroma and adipose tissue, with little to no invasive carcinoma.

To examine the biological basis of CTX-$\tau$ scores, we next investigated molecular correlates (\textbf{\Cref{fig:explainability}c--d}). 
These analyses were conducted in TCGA (n = 348), the only cohort in our study with transcriptome-wide RNA-sequencing paired to histopathology.
As this cohort was part of the development set, these analyses are exploratory and hypothesis-generating.
CTX-$\tau$ high-benefit tumors exhibited multiple mechanisms that may drive chemosensitivity.
They were enriched for proliferative signaling pathways (MYC targets, mTOR signaling), cell cycle and division (E2F targets, G2M checkpoint, mitotic spindle), and DNA damage response pathways, consistent with replication stress arising from accelerated proliferation.
To further interrogate the linkage with DNA damage, we examined a genomic signature of homologous recombination deficiency, finding this was positively associated with CTX-$\tau$.
Meanwhile, immune activation pathways (IFN-$\alpha$ and IFN-$\gamma$ response) were also up-regulated in CTX-$\tau$ high-benefit tumors, which may mediate chemotherapy-induced immunogenic cell death.
To test a potential mechanistic link between cytosolic DNA and immune activation~\citep{li2021metastasis}, we specifically examined enrichment of the cGAS--STING pathway, finding that it was also enriched in CTX-$\tau$ high-benefit tumors.
By contrast, CTX-$\tau$ low-benefit tumors exhibited up-regulation of epithelial mesenchymal transition (EMT) and angiogenic pathways --- which may serve as mediators of microenvironment-mediated drug resistance~\citep{meads2009environment}. These tumors were also enriched for early estrogen response, consistent with preferential sensitivity to ET over CET.

In summary, we identified concordant morphological and molecular correlates of CTX-$\tau$ predictions, indicating that model-detected chemosensitivity could be grounded in interpretable tumor biology.
CTX-$\tau$ high-benefit tumors harbored molecular programs of proliferation and cell cycle progression, and replication stress with homologous recombination deficiency --- hallmarks of susceptibility to chemotherapy-induced cytotoxicity --- consistent with their high cellularity and identifiable mitotic figures observed on histology.
Conversely, CTX-$\tau$ low-benefit tumors were enriched for estrogen response signatures, consistent with preferential endocrine sensitivity, alongside EMT and angiogenesis pathways that may be contribute to chemo-resistance --- mirroring their stroma-rich, hypocellular appearance. 
Together, these biologically plausible correlates may help support clinician trust and adoption of CTX.

\begin{figure}[h]
    \centering
    \includegraphics[width=1.0\linewidth]{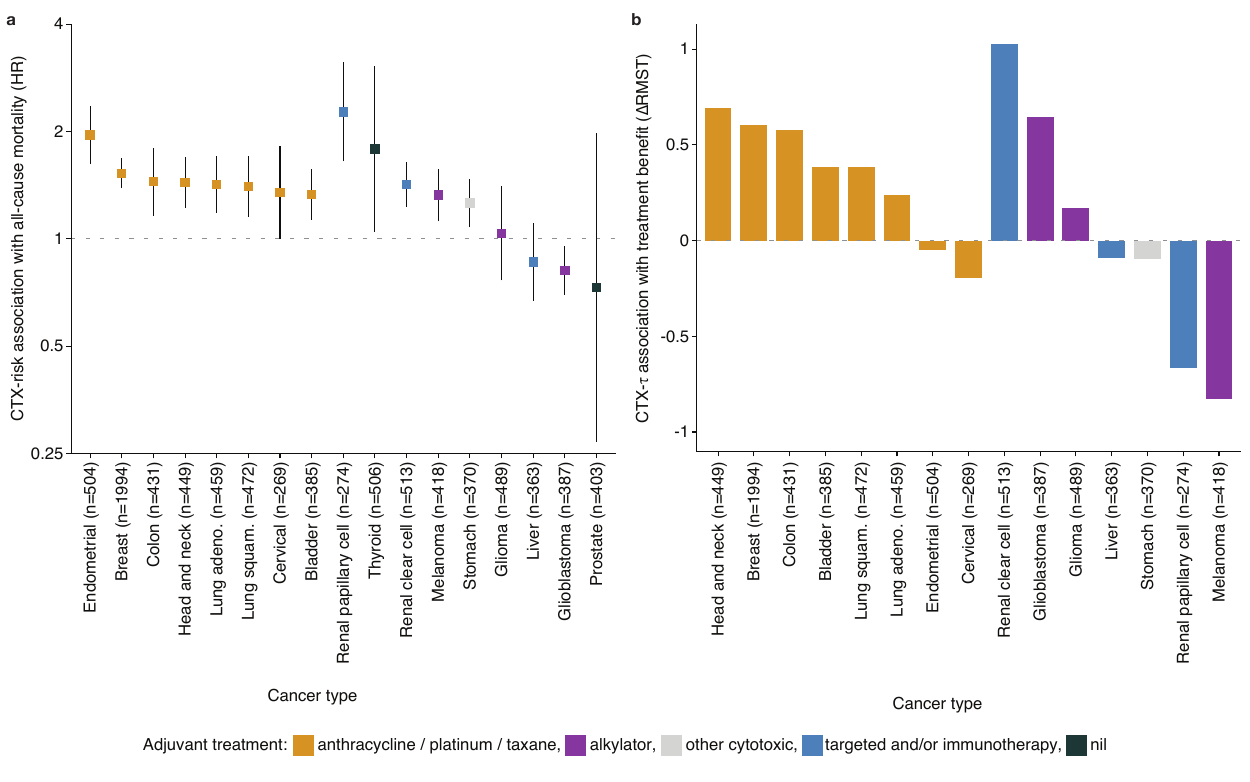}
    \caption{\textbf{Zero-shot transfer of CTX prognosis and chemosensitivity predictions across cancer types.}
    The frozen CTX model, trained only on breast cancer, was applied without fine-tuning or re-calibration to patients spanning 16 cancer types from The Cancer Genome Atlas (TCGA), using overall survival as the endpoint. Performance of CTX in the evaluation set is also displayed (as ``breast'') for reference. Each cancer type is colored by adjuvant systemic therapy class --- anthracycline/platinum/taxane (gold), alkylator (purple), or other cytotoxic (light grey); targeted and/or immunotherapy (blue); or none (green). CTX was trained to predict benefit from anthracycline-, platinum- and/or taxane-based regimens, therefore prognostic and predictive transfer is expected primarily in cancer types treated with mechanistically comparable cytotoxic regimens.
    \textbf{a}, Box plot shows prognostic transfer of CTX-risk, assessed independently in each cancer type by univariable Cox regression (n = 6,692 patients across 16 cancer types, and n = 1,994 CTX breast evaluation set). Points show the hazard ratio for CTX-risk per cancer type; vertical bars are 95\% CIs. The horizontal dotted line marks the null (hazard ratio = 1).
    \textbf{b}, Bar plot shows predictive transfer of CTX-$\tau$, evaluated in the cancer types routinely treated with adjuvant systemic therapy (n = 5,783 patients across 16 cancer types, and n = 1,994 CTX breast evaluation set). Within each cancer type, patients were ranked by CTX-$\tau$ and split into high- and low-benefit subgroups at the 90th percentile; in an IPW-adjusted cohort we computed the difference in restricted mean survival time (RMST) between chemotherapy-treated and untreated patients within each subgroup. Each bar is the high-benefit minus low-benefit $\Delta$RMST (years). Positive $\Delta$RMST values indicate that the subgroup CTX-$\tau$ predicts will benefit most gained more survival time from chemotherapy than the low-benefit subgroup (correct predictive ordering); negative values indicate the reverse.
    }
    \label{fig:pancancer}
\end{figure}

\subsection{Zero-shot transfer of CTX prognosis and chemosensitivity predictions across cancer types} \label{sec:results:pancancer}

To assess whether CTX may provide a universal strategy for predicting treatment outcomes, we tested whether its prognostic and chemosensitivity predictions generalize beyond breast cancer.
We used a strict zero-shot framework~\citep{radford2021learning}: the frozen model was applied directly to 6,692 patients spanning 16 cancer types from TCGA, with no fine-tuning or re-calibration. 
Performance of CTX in the held-out breast evaluation set (n = 1,994) was used as a reference.
Overall survival (OS) served as the endpoint, as it is the most completely annotated outcome in TCGA. 
We stratified results by adjuvant treatment class, reasoning that the anthracycline-, platinum- and taxane-based regimens standard in HR+/HER2$-$ breast cancer share mechanisms with the cytotoxic regimens used in some other malignancies.

We first tested whether the model's baseline prognostic prediction (CTX-risk), associated with OS across cancer types. We used baseline risk here to isolate the prognostic signal from any off-target treatment effects.
In each cancer type, CTX-risk was included as the sole covariate in a Cox model, with significance assessed by likelihood ratio test against the null model.
Higher CTX-risk was associated with worse OS in 11 of 16 (69\%) cancer types, indicating that the prognostic signal learned from breast cancer histopathology transfers broadly across malignancies (\textbf{\Cref{fig:pancancer}a}). 
Transfer was strongest in tumors biologically similar to HR+/HER2$-$ breast cancer --- endometrial carcinoma, which shares estrogen-signalling biology, and other epithelial adenocarcinomas (colon, lung, head and neck) --- and weakest in the tumors most distant from a breast epithelial phenotype, such as central nervous system tumors (glioma and glioblastoma).
Consistent with a shared cytotoxic axis, 6 of 7 (86\%) cancer types treated with anthracycline-, platinum- and/or taxane-based regimens showed significant prognostic transfer.

Next, we tested whether the chemosensitivity signal of CTX-$\tau$ transferred. 
For each of the 14 cancer types routinely treated with adjuvant systemic therapy (n = 5,783), we ranked patients by CTX-$\tau$, split them into high- and low-benefit subgroups, and compared the survival gain from chemotherapy between subgroups using an IPW-adjusted difference in restricted mean survival time ($\Delta$RMST).
CTX-$\tau$ correctly ordered chemotherapy benefit in 8 out of 14 (57\%) cancer types (\textbf{\Cref{fig:pancancer}b}).
The signal transferred most reliably where we anticipated: in 5 of 7 (71\%) cancer types treated with anthracycline-, platinum- and/or taxane-based regimens --- head and neck, colon, bladder, and lung squamous and adenocarcinoma --- the CTX-$\tau$ high-benefit subgroup gained more survival time from chemotherapy than the low-benefit subgroup.
Transfer failed predominantly in cancers treated with mechanistically unrelated regimens: alkylator therapy in glioblastoma, glioma, and melanoma; targeted or immunotherapies in renal and hepatocellular carcinoma. 

A frozen, breast-trained causal AI model recovered prognostic ordering in most cancer types and correctly ordered treatment benefit in those treated with mechanistically related regimens. This indicates that CTX has learned generalizable, treatment-specific patterns of tumor aggressiveness and chemosensitivity~\citep{bommasani2021,moor2023generalist}.

\section{Discussion}

Precision oncology aims to match the right treatment to the right patient~\citep{subbiah2022accelerated}. 
Realizing this vision requires advancing from population-level estimates to predictions personalized to the individual~\citep{kent2018personalized}. 
Yet, biomarkers are conventionally designed and evaluated in coarse subgroups, rather than for the individual patient. 
Existing biomarker discovery approaches, including recent AI models, estimate clinical outcomes regardless of treatment and are therefore prognostic rather than predictive~\citep{mcdonald2023computational, wang2026discovery}. 
To guide the selection of the optimal treatment, a predictive biomarker should instead estimate the benefit of a specific therapy for a specific patient~\citep{ballman2015biomarker, aldea2025esmo}.
Causal AI models offer a route to personalized estimation of treatment benefit~\citep{feuerriegel_natmed_2024}. 
To date, however, causal approaches in oncology have shown limited efficacy, plausibly reflecting small cohorts or insufficiently expressive representations~\citep{weberpals2025opportunities}.
A pathology foundation model, pretrained on a large-scale histology dataset, may address both limitations by encoding an information-dense representation of tumor phenotype, thereby reducing the data burden on the causal model itself.

Overtreatment is a central problem in early breast cancer --- for many patients, the absolute survival benefit of adjuvant chemotherapy is small~\citep{Sparano_NEJM_2018, Kalinsky_NEJM_2021} and is outweighed by treatment-related toxicity~\citep{Curigliano_TheBreast_2016}.
Clinical guidelines advocate for the use of prognostic tests, such as the 21-, 50- and 70-gene assays, as a level 1 indication for the consideration of adjuvant systemic therapy~\citep{NCCN_BreastCancer_2026}. 
While the increased adoption of prognostic tests for chemotherapy decision-making has partially reduced overtreatment~\citep{katz_jama_2018}, this approach has a critical flaw --- it tacitly assumes that a patient at high predicted risk of recurrence will benefit from chemotherapy.
However, prognosis and treatment benefit are distinct qualities: a prognostic marker reflects the natural history of disease irrespective of therapy, whereas a predictive marker identifies which patients benefit from a specific treatment~\citep{ballman2015biomarker, aldea2025esmo}.
A strong prognostic marker might not predict differential chemotherapy benefit~\citep{clark2008prognostic} --- what is needed instead are markers that estimate this benefit directly to guide patient-tailored therapy~\citep{oldenhuis2008prognostic}.
These prognostic tests also require molecular testing, incurring cost and tissue demands that limit universal access~\citep{mateo2022delivering}.

Here, we developed CTX to meet this need, estimating personalized chemotherapy benefit in HR+/HER2$-$ early breast cancer directly from routine histopathology and clinical data.
To our knowledge, this is the first instance of applying causal multi-modal AI --- chaining counterfactual regression with an image-based foundation model --- for predicting treatment outcomes~\citep{feuerriegel_natmed_2024}.

CTX was developed on 9,141 patients and evaluated on a held-out set of 1,994 patients drawn from independent cohorts.
This scale was necessary to learn a rich causal representation to support estimation of chemosensitivity.
Applying causal AI to predict personalized chemosensitivity required several extensions beyond the standard setting. First, the endpoint is a censored time-to-event outcome rather than a binary label, so treatment effects must be estimated in a survival framework. Second, cohorts with differing treatment-assignment mechanisms and baseline recurrence rates must be integrated into a single predictive framework. Third, the conditioning variables extend beyond structured clinical covariates to high-dimensional morphological features extracted from routine histology by a pathology foundation model. Together, these elements ground the prediction task in both clinicopathological factors and tumor phenotype, with the aim of identifying the patients most likely to benefit from chemotherapy.

CTX-$\tau$ met the formal criterion for a predictive biomarker, exhibiting a significant treatment-by-biomarker interaction, whether used as a binary or continuous predictor~\citep{ballman2015biomarker}.
In addition, CTX-prognostic showed strong discrimination, near-perfect calibration, and remained an independent predictor of recurrence after adjustment for standard clinicopathological variables. 
Performance held across both 5- and 10-year horizons, a key consideration given the persistent hazard of late recurrence in HR+/HER2$-$ disease~\citep{pan201720}.
These analyses demonstrate that causal AI can advance treatment selection in early breast cancer beyond population-averaged prognostic estimates to personalized chemosensitivity predictions.

For a model to achieve clinical adoption, it must use routinely available data, address an actionable decision point, and demonstrate better performance than existing tests~\citep{markowetz2024all}.
CTX satisfies all three criteria: it relies solely on H\&E morphology and clinical data; it addresses a concrete decision point --- whether to recommend adjuvant chemotherapy; and it produces an output --- personalized chemosensitivity --- that maps directly onto that decision. 
The treatment-by-biomarker interaction for CTX was the strongest among all evaluated predictors, exceeding both the Oncotype DX recurrence score --- the most widely used multi-gene assay in current practice --- and standard clinicopathological variables.
Model explainability is another prerequisite for adoption. 
We therefore linked CTX predictions to morphological and molecular features, finding that CTX-$\tau$ high-benefit tumors couple proliferation and DNA-damage-response programs with homologous-recombination deficiency and a primed immune microenvironment --- a state mechanistically consistent with susceptibility to chemotherapy-induced cytotoxicity. 
Grounding an AI model in mechanisms of this kind may ease its path to clinical adoption~\citep{topalian_nrc_2016}.

We discovered that CTX separates two quantities that existing assays conflate --- recurrence risk and chemosensitivity --- capturing treatment effect heterogeneity that prognostic tests miss. 
Moreover, population-based estimates obscure clinically meaningful variation that CTX recovers at the individual level, distinguishing patients who share a clinicopathological profile yet differ in predicted benefit.
Decision analyses indicate that model-guided chemotherapy allocation could reduce overtreatment relative to historical prescribing patterns, while directing chemotherapy toward the patients most likely to benefit.

Some limitations temper the conclusions of this study. 
First, evaluating any model of treatment benefit is constrained by the fundamental problem of causal inference: counterfactual outcomes are never observed~\citep{holland1986statistics}. 
All cohorts in this study were observational with respect to chemotherapy. 
We mitigated confounding through representation balancing during training and inverse propensity weighting at evaluation, but residual confounding from unmeasured factors cannot be excluded.
Our companion study evaluates CTX in TAILORx~\citep{Sparano_NEJM_2018}, a phase III trial that randomized patients at intermediate genomic risk to ET or CET, quantifying the accuracy of prognostic and chemotherapy benefit predictions benchmarked relative to Oncotype DX~\citep{chan_tailorx_2026}.
Second, CTX models treatment as a binary contrast between CET and ET; the choice of chemotherapy regimen, dose, and schedule is not resolved and represents a natural extension. 
Third, the mechanistic associations we report are correlative; the proposed links between the biology of CTX-$\tau$ high-benefit tumors and chemosensitivity are a hypothesis requiring prospective and experimental validation.

Future analyses may extend our causal multi-modal AI approach toward a general strategy for predicting treatment outcomes across cancer types.
CTX was trained solely on breast cancer, yet the model's prognostic and chemosensitivity predictions transferred to other cancer types, with no per-cancer fitting. 
This zero-shot transfer distinguishes our result from prior pan-cancer analyses of AI pathology models, which reported performance by fitting or fine-tuning a separate prognostic head for each cancer type~\citep{wang_nature_2024, xiang2025vision}. That our breast-trained causal model identifies risk and chemosensitivity in the majority of cancer types is a substantially stronger form of generalization.
These data suggest CTX may capture morphological determinants of chemosensitivity that are shared across malignancies, motivating further cross-cancer validation. 
Recent scrutiny of cancer-drug approvals has cautioned against surrogate endpoints, such as objective response rate, that often fail to predict how long patients actually live~\citep{shkabari_jama_2026}. 
Our framework aligns with this renewed emphasis on survival by estimating treatment effects on a time-to-event outcome directly.
Future work may also improve the model via enrichment with orthogonal information such as spatially resolved molecular data~\citep{aung_natgenetics_2025}, and longitudinal signals such as circulating tumor DNA~\citep{wang_cell_2019}.  
By learning such joint embeddings, our causal multi-modal AI framework could in principle be extended to learn out-of-domain effects for other therapies and cancer types with minimal additional data --- serving as a causal ``world model'' for oncology.

\newpage
\section{Methods} \label{sec:methods}

\subsection{Data}

We assembled 17 cohorts of patients with early breast cancer from 10 countries.
Cohorts were assigned in their entirety either to model development (12 cohorts, n = 9,141) or to evaluation (5 cohorts, n = 1,994).
This study was conducted in accordance with the ethical principles of the Declaration of Helsinki and applicable regulatory requirements, including ICH-GCP, relevant US/Canadian human research protection regulations, and TCPS2, as applicable. Informed consent was waived due to the minimal risk to the subjects, the retrospective collection of data, and the use of only de-identified data.

For each patient, we collected at least one de-identified H\&E whole slide image (WSI). Standard clinicopathological features --- age, tumor and nodal stage, ER, PR and HER2 status, and histological subtype --- were harmonized across all cohorts. 
The data also includes an AI genomic risk score, a continuous prediction of the research-based genomic risk score (\textbf{\Cref{sec:research_odx}}).
All patients in the development cohorts were female, and had early-stage disease, known adjuvant chemotherapy status, an annotated clinical outcome, and no prior breast cancer.
They were deliberately not restricted by receptor subtype or treatment, so that the model could learn from the full morphological and prognostic range of early breast cancer.
Evaluation cohorts were restricted to the intended deployment population: HR+/HER2$-$ early invasive disease, receipt of adjuvant endocrine therapy (ET) or chemoendocrine therapy (CET), and no adjuvant targeted therapy (\textbf{Extended Data \Cref{fig:exclusion_diagrams}}).
Chemotherapy was not randomized in any cohort. All cohorts were observational except UNIRAD, a phase III trial randomizing everolimus; only its control arm, consistent with the no-targeted-therapy criterion, is included. 

The primary endpoint was recurrence-free interval (RFI), defined as the time from diagnosis to the first invasive local, regional or distant recurrence.
This is modeled on the STEEP v2.0 system~\cite{tolaney_jco_2021}, except deaths were treated as censoring events, so that the training target isolated the five-year recurrence-free indicator.

\subsection{Pathology foundation model} \label{sec:methods:falcon}
H\&E-derived morphological features were extracted with Falcon, successor to Kestrel~\citep{cappadona_neuripsworkshop_2024}, our pathology foundation model. Falcon is a ViT-Giant (1.1 billion parameters) pretrained via self-supervised learning (DINOv2~\citep{oquab2024dino}) on over two billion H\&E patches from 180,000 whole-slide images, and achieves state-of-the-art performance across patch- and slide-level benchmarks~\citep{cappadona_sabcs_falcon}. Pretraining used no outcome labels, and no slides used for pretraining were used in the training or evaluation of CTX. Falcon was frozen throughout training and evaluation of CTX.

Slides were processed identically during development and evaluation. Tissue regions were identified with a custom DeepLabV3~\citep{chen2017rethinking} segmentation model and divided into non-overlapping 224 $\times$ 224 pixel patches at 0.5\,\textmu m-per-pixel (approximately 20$\times$ magnification), each encoded by Falcon into a 1,536-dimensional embedding.

\subsection{Multi-modal architecture} \label{sec:methods:multimodal}
Patch embeddings extracted by Falcon were aggregated into a slide-level representation by a gated-attention multiple instance learning layer~\citep{pmlr-v80-ilse18a}. In parallel, clinical covariates --- tumor and nodal stage, patient age, ER/PR/HER2 receptor status, ductal and lobular histology, and an AI-derived genomic risk score (\textbf{\cref{sec:research_odx}}) --- were encoded by a tabular ResNet~\citep{gorishniy2021revisiting}. The two representations were fused by gated attention into a single multi-modal representation, which is used as the input to the model's treatment-specific output heads.

\subsection{Discrete-time survival modeling} \label{sec:methods:survival}

The primary endpoint, RFI, is a censored time-to-event outcome, so CTX predicts a survival curve across follow-up times rather than a single probability. 
We adopted a discrete-time parameterization: follow-up is partitioned into a fixed set of intervals, and for each interval the model predicts a hazard representing the probability of recurrence within that interval, given that the patient was recurrence-free at its start \citep{cox1972regression, kvamme2021continuous, gensheimer2019scalable}. 
Interval boundaries were derived from evenly spaced quantiles of the event durations in the development set, then piecewise-linearly rescaled so that one boundary fell at exactly five years. 
The recurrence-free probability at five years is calculated as the cumulative product of one minus the hazards over the intervals up to and including the five-year boundary.
The training loss is a binary cross-entropy on the predicted hazards, evaluated only over the intervals in which a patient remained at risk. 
A patient censored before five years therefore contributes a term for every interval they were observed through and none thereafter, so right-censoring is handled without imputing unobserved event times.
The number of intervals was tuned by hyperparameter search (\textbf{\cref{sec:methods:multimodal_architecture}}, \textbf{Extended Data~\cref{tab:hyperparameters}}).

\subsection{Causal modeling} \label{sec:methods:causal}
For every patient, only one outcome is observed under the treatment they received, yet a benefit estimate requires a prediction under both treatment assignments. 
Therefore, the model passes the fused multi-modal representation to two treatment-specific output heads, one for CET and one for ET, following the family of counterfactual regression models (CFRNet)~\cite{johansson_icml_2016,shalit_icml_2017}. 
Each head emits the discrete-time hazards described in\textbf{~\cref{sec:methods:survival}}, so every patient receives a full survival curve under each treatment assignment.

The multi-modal backbone and the per-treatment output heads are therefore trained on different data: the multi-modal representation is learned from all patients, whereas each head is fitted only on the patients who received the corresponding treatment. 
Because chemotherapy was not randomly assigned, the multi-modal representations of ET and CET patients may have different distributions. 
Therefore, to better align these distributions and encourage more reliable counterfactual prediction, we applied a maximum mean discrepancy penalty to reduce the distributional shift between the treatment groups in the representation space~\citep{gretton2012kernel,shalit_icml_2017}.

The two per-treatment predictions from the CFRNet define CTX's outputs.
CTX-prognostic is the predicted five-year recurrence probability under the treatment a patient actually received --- baseline risk for ET patients, residual risk for CET patients.
CTX-$\tau$ is the difference between the two heads' five-year predictions: the gain in five-year recurrence-free probability from adding chemotherapy, with positive values denoting predicted benefit from CET.
Patients with CTX-$\tau$ greater than or equal to 2 percentage points were classified as high-benefit from chemotherapy.
This threshold was informed by a survey of patient preferences~\citep{duric2005patients} and the 2025 St Gallen International Breast Cancer Consensus Statement on individualizing therapy for patients with early breast cancer~\citep{burstein2025tailoring}.

\subsection{Training methodology}

The training objective of the model comprises three terms. 
First, the discrete-time survival loss of\textbf{~\cref{sec:methods:survival}}, evaluated only at the head corresponding to each patient's observed treatment; no part of the loss requires an outcome that was never observed.
Second, a penalty on the discrepancy between the two treatment groups' representation distributions, measured by maximum mean discrepancy~\cite{gretton2012kernel}.
This reduces distributional distance (in the embedding space) between the two treatment arms, encouraging a representation under which outcomes can be predicted accurately for either arm.
Third, an elastic-net penalty on the network parameters.
The full objective is discussed in\textbf{~\cref{sec:methods:objective}}.

CTX was trained under three-fold leave-one-cohort-out multiple-source cross-validation~\cite{geras_pmlr_2013} (see\textbf{~\cref{sec:methods:mscv}} for more details). 
Three development cohorts (NYU, Cardiff, Sydney) were designated as held-out ``validation'' datasets to use for hyperparameter selection, early stopping and calibration.
Each was held out in turn, while the remaining 11 development cohorts were used for training.
The ten hyperparameter configurations with the best validation performance in each validation dataset were selected, and the corresponding model predictions were ensembled by uniform averaging.
Beta calibration~\cite{kull_betacal} was then applied to the ensemble's factual five-year predictions. 
This yielded three sets of calibrated predictions (one for each validation dataset), which were then averaged to produce the final predictions.
Architecture, objective, and hyperparameter search space are given in\textbf{~\cref{sec:methods:multimodal_architecture}}, \textbf{\cref{sec:methods:objective}}, and \textbf{Extended Data~\cref{tab:hyperparameters}}.

\subsection{Adjustment for confounding and propensity score estimation}

Evaluating whether CTX-$\tau$, or any biomarker, identifies who benefits from chemotherapy requires adjusting for confounding, because chemotherapy was not randomly assigned in the evaluation data.
All between-arm treatment contrasts were therefore adjusted by inverse propensity weighting~\citep{austin2011introduction}. 
The propensity score --- each patient's probability of receiving chemoendocrine therapy given their clinical covariates --- was estimated by gradient boosting with default hyperparameters~\cite{prokhorenkova2018catboost} and five-fold cross-fitting~\cite{kennedy2024semiparametric}.
This yielded out-of-fold predictions for every patient, which were then calibrated by beta calibration~\citep{kull_betacal}. 
The weight assigned to each patient is the inverse of the estimated probability of the treatment they actually received, clipped at 10 to limit variance.
Weighting reduced the mean absolute standardized mean difference across covariates from 0.376 to 0.125 and the maximum from 1.036 to 0.281.

Further description of propensity estimation and balance diagnostics is in\textbf{~\cref{sec:methods:eval:propensities}}.

\subsection{Evaluation}

The five evaluation cohorts were pooled into a single dataset and analyses were performed on the pooled dataset.

The principal analysis of chemotherapy-benefit prediction was the interaction between dichotomized CTX-$\tau$ and chemotherapy receipt in an IPW-adjusted Cox model with robust standard errors (sandwich variance estimator), using the 2\% high-benefit threshold.
Further analyses repeated the test with CTX-$\tau$ as a continuous score (reported per standard deviation), and in an unweighted model adjusted for age, T- and N-stage, grade and histology.
Within each stratum, the CET-versus-ET contrast was summarized in three ways. First, as the hazard ratio from an IPW-adjusted Cox model.
Second, as the difference in five-year recurrence-free probability between IPW-adjusted Kaplan--Meier curves. 
Finally, as the difference in five-year restricted mean survival time between the same curves.
Stratum-level significance was assessed by the score test from the IPW-adjusted Cox model.

To place CTX-$\tau$ in the context of existing biomarkers, we benchmarked it against a panel of clinical, pathological, and molecular baselines. Each biomarker was evaluated by its interaction with chemotherapy receipt in an IPW-adjusted Cox model --- the same test applied to CTX-$\tau$ --- with interaction effects reported per standard deviation. As availability varied across comparators, each was assessed in the subset of patients for whom it was recorded: age in all 1,994 patients; tumor stage in 1,971; nodal stage in 1,982; tumor grade in 1,520; histological subtype in 1,526; Ki67 in 229; and Oncotype DX, the principal comparator, in 983 patients.

Prognostic performance was assessed for CTX-prognostic.
Discrimination was measured by time-dependent area under the receiver operating characteristic curve, the association with recurrence by the hazard ratio from multivariable Cox regression adjusted for age and T- and N-stage, and calibration by comparing Kaplan--Meier-estimated observed event probabilities within quintiles of predicted risk against mean predicted risk.

The decision-analytic analysis in \textbf{\Cref{sec:results:decision}} estimated the five-year recurrence-free rate under treatment-allocation strategies assigning chemotherapy to the highest $q$ percent of patients according to CTX-$\tau$ using an IPW-adjusted Kaplan--Meier estimator. 
The recurrence-free rate was estimated for every 10 percentage point increase in the chemotherapy rate, from 0\% to 100\% of patients receiving CET.
The results were estimated over the full range of treatment proportions but displayed up to 60\%, within one standard deviation of the mean observed treatment rate.
Methods implementing the decision-support, explainability, molecular-correlate and generalizability analyses are given in \textbf{\cref{sec:methods:eval:decision}}, \textbf{\cref{sec:methods:eval:explainability}}, \textbf{\cref{sec:methods:eval:correlates}}, and \textbf{\cref{sec:methods:eval:pancancer}}. The molecular analyses are exploratory as they were performed in TCGA-BRCA, a development cohort.

\subsection{Statistical analysis}
All tests were two-sided unless otherwise stated.
The numbers of patients included in each analysis are annotated in the corresponding figures.
Confidence intervals are reported as 95\% CI = [lower, upper] throughout.
As the data were collected retrospectively, no statistical methods were used to predetermine sample size.
Statistical analyses were performed in Python (version 3.10) or R (version 4.3.3). Hazard ratios and p-values from Cox proportional hazards regression were calculated using the \texttt{survival} package, through univariable or multivariable analyses as stated. Kaplan--Meier survival curves were plotted using the \texttt{survminer} and \texttt{ggsurvfit} packages; score tests from IPW-adjusted Cox models were obtained from \texttt{coxph} in the \texttt{survival} package. Time-dependent area under the receiver operating characteristic curve was computed using the \texttt{tdROC} package~\citep{tdroc} and concordance indices~\citep{harrell1996multivariable} were obtained from the \texttt{concordance} function in the \texttt{survival} package. Bootstrap 95\% confidence intervals for the time-dependent AUC and the concordance index were derived from 1,000 replicates. Tests involving correlations were performed using Spearman's or Pearson's method, as specified. Comparisons of continuous variables between two groups were performed using the Wilcoxon rank-sum test, and comparisons across more than two groups were performed using the Kruskal--Wallis test.
Hardware is described in\textbf{~\cref{sec:methods:mscv}}.
Investigators were not blinded to cohort identity during model development. The pathologists who annotated morphological clusters were blinded to model predictions and clinical outcomes.
No pre-specified analysis plan was registered, and the evaluation cohorts were held out from training and model selection but were not sequestered from analysis.

\subsection{Data availability}

The datasets analyzed in this study were obtained from 17 cohorts under data use agreements that in most cases preclude public redistribution of individual-level data. Requests for access to each cohort should be directed to the respective data custodians, subject to the terms of the original agreements and applicable ethical and legal frameworks.
Publicly available cohorts can be accessed directly. 
TCGA data, including diagnostic H\&E whole-slide images, are available from the NCI Genomic Data Commons (\url{https://portal.gdc.cancer.gov}). 
METABRIC molecular and clinical data are available via cBioPortal (\url{https://www.cbioportal.org}).

\subsection{Code availability}

The full codebase underlying the AI model is proprietary and cannot be publicly released. The model and selected components of the code may be made available to academic researchers for non-commercial research purposes, subject to institutional approval and execution of appropriate data use and/or material transfer agreements. Requests will be evaluated on a case-by-case basis to ensure compliance with data privacy, intellectual property, and commercial considerations. To facilitate reproducibility, we provide detailed methodological descriptions in the Methods section, including model architectures, training procedures, and hyperparameter selection strategies, and we reference all open-source libraries and frameworks used in the implementation. In addition, the code used to generate the main figures and aggregate results presented in this study can be made available upon reasonable request.

\section*{Acknowledgments}
This study makes use of data generated by the Molecular Taxonomy of Breast Cancer International Consortium (METABRIC); funding for the project was provided by Cancer Research UK and the British Columbia Cancer Agency Branch~\citep{curtis2012genomic}.
The results published here are in part based upon data generated by the TCGA Research Network (\url{https://www.cancer.gov/tcga}).
Tissues and samples were received from the Australian Breast Cancer Tissue Bank, supported by the National Health and Medical Research Council of Australia, the Cancer Institute NSW, and the National Breast Cancer Foundation, and made available to researchers on a non-exclusive basis.
Biosamples were obtained from the Wales Cancer Bank~\citep{parry2018wales}, funded by Health and Care Research Wales; other investigators may have received specimens from the same subjects.
Biological materials were provided by the Ontario Tumour Bank, supported by the Ontario Institute for Cancer Research through funding from the Government of Ontario; the views expressed are those of the authors and do not necessarily reflect those of the Government of Ontario.
We acknowledge the roles of the Breast Cancer Now Tissue Bank in collecting and making available the samples and/or data, and the patients who generously donated their tissues and shared their data for use in this publication.
We acknowledge Pieter Westenend and Stichting Albert Schweitzer Ziekenhuis, Laboratorium voor Pathologie, as providers of data from Dordrecht.
We acknowledge the University of Chicago for providing the UChicago cohort~\citep{howard2023integration}.

\section*{Author contributions}
DB, JB, JW and KJG contributed to study conception and design.
JB, AM, LB, JP, KGZ, JC and KJG contributed to the design of the models used in this study.
DB, JB, AM, LB, JP, KGZ, JC, CT, CL and KJG contributed to the evaluation methods used in this study.
YW, HS, RB, SK, TK, DP, CB, PW, PHC, FD, FA, FMLP, TB, FH, FJE, KK, LP and JW provided clinical guidance.
VS, BP, and SC provided statistical and informatics guidance.
JB, AM, LB, JP, KGZ and JC wrote code, developed infrastructure and trained models throughout the study.
BM, CM and JL worked on data preparation.
DB, AM, LB, JP, KGZ, CT and CL performed evaluation and analysis.
DB, JB, AM, LB, JP, KGZ, JC, CT and KJG worked on drafting and revising the manuscript.
All authors critically reviewed the paper and the results and approved the final version.

\section*{Competing interests}
DB, JB, AM, LB, JP, KGZ, JC, CT, CL, BM, FJE, LP, JW and KJG are equity holders of Ataraxis AI. JB, JP, KGZ, JC, LB, AM, DB, CL, JW and KJG are inventors on a US patent application filed corresponding to some of the methodological aspects of this work. The remaining authors declare no competing interests.

\newpage
\bibliographystyle{unsrtnat} 
\bibliography{references} 

\newpage
\section{Extended Data}
\setcounter{section}{0}

\setcounter{figure}{0}
\renewcommand{\figurename}{Extended Data Fig.}
\renewcommand{\thefigure}{\arabic{figure}}

\begin{figure}[h]
    \centering
    \includegraphics[width=1.0\linewidth]{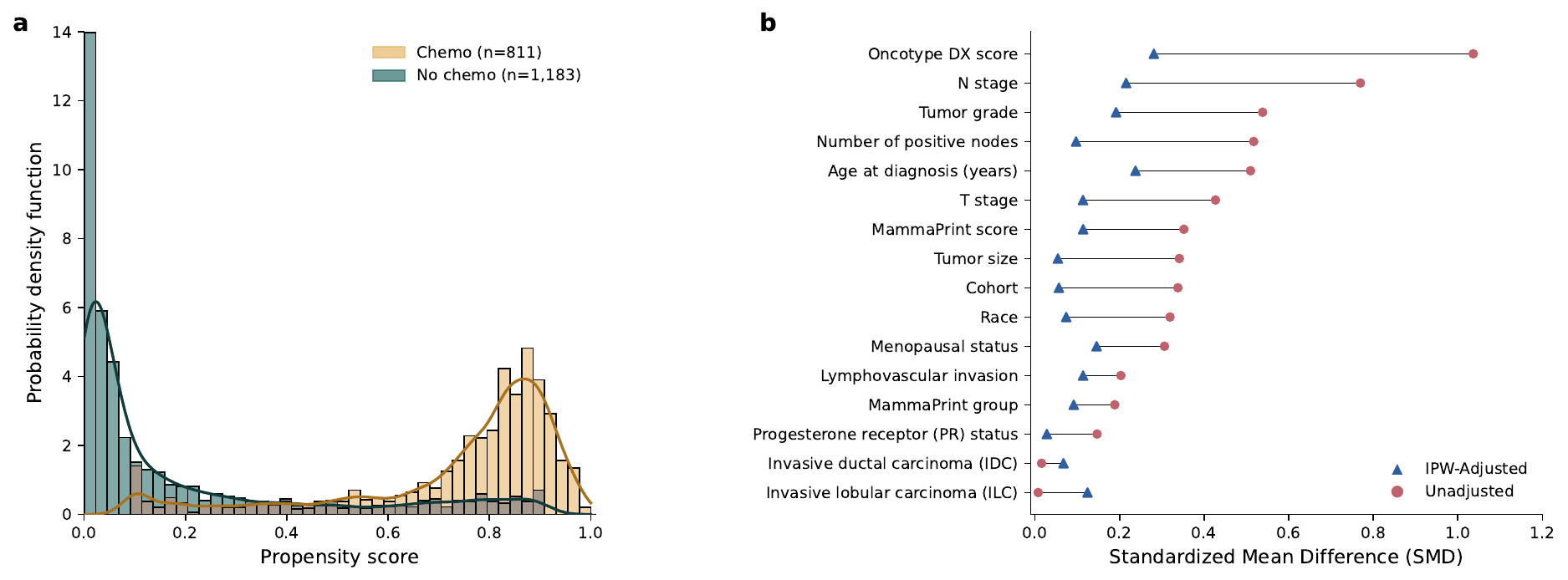}
    \caption{
    \scriptsize
    \textbf{Propensity adjustment.}
    \textbf{a}, Histogram of propensity score estimates colored by treatment status. Scores are clipped so that no patient has an inverse propensity weight (IPW) greater than 10.
    \textbf{b}, Standardized mean differences between chemoendocrine and endocrine therapy patients, before and after IPW-adjustment.
    }
    \label{fig:propensity_scores}
\end{figure}

\clearpage
\begin{figure}[!ht]
    \centering
    \includegraphics[height=0.85\textheight,keepaspectratio]{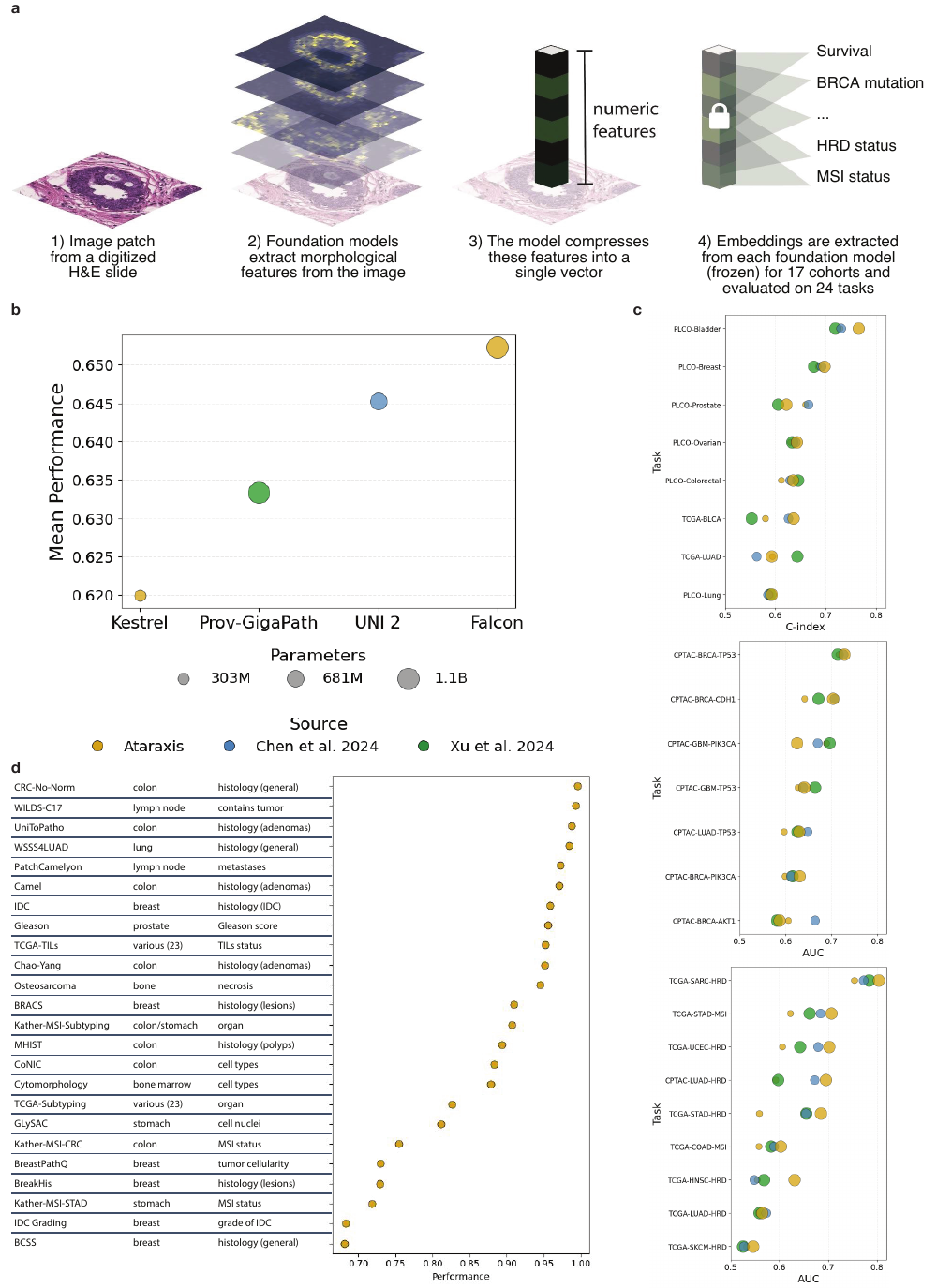}
    \caption{
    \scriptsize
    \textbf{Pathology foundation model benchmarking.}
    The pathology foundation model used by CTX (\textbf{\Cref{fig:figure_1}d}), Falcon, is a ViT-Giant (1.1B parameters) pretrained with DINOv2~\citep{oquab2024dino} on over two billion H\&E patches from 180,000 whole-slide images. 
    Falcon was benchmarked on a slide-level task suite against its predecessor Kestrel~\citep{cappadona_neuripsworkshop_2024} (ViT-Large, 303M parameters) and two publicly available pathology foundation models, UNI 2~\citep{chen_naturemed_2024} (681M parameters) and Prov-GigaPath~\citep{xu_nature_2024} (1.1B parameters).
    No pretraining slide was used in any benchmark task, and no benchmark task overlapped with any datasets used for evaluating CTX.
    \textbf{a}, Benchmarking protocol. Each foundation model extracts a patch-level embedding for all patches in each slide. All embeddings from a slide are mean-pooled, and this mean-pooled embedding is used as a feature vector to evaluate the performance of the foundation model on survival or classification tasks.
    \textbf{b}, Aggregate performance on a suite of 24 slide-level histopathology evaluation tasks. Performance is the C-index for survival tasks and AUC for mutation and biomarker tasks. Falcon achieved the highest mean performance across all tasks. The gain over Kestrel, trained using similar protocol on an order of magnitude fewer data, indicates that generalization performance scales with model capacity and training duration.
    \textbf{c}, Per-task slide-level performance underlying \textbf{b}. Axis labels describe the cohort, and the target for classification tasks; in the case of survival tasks, the target was disease-specific survival. 
    \textbf{d}, Performance of Falcon on a suite of 24 patch-level histopathology evaluation tasks. Hyperparameters for Falcon were tuned by optimizing performance on this task suite. Each task tests the model's ability to identify a specific property of cancerous tissue. All tasks are classification tasks, except for BreastPathQ and CoNIC which are regression tasks. The performance of the final Falcon model is shown.
    }
    \label{fig:falcon}
\end{figure}

\clearpage
\begin{figure}[h]
    \centering
    \includegraphics[width=1.0\linewidth]{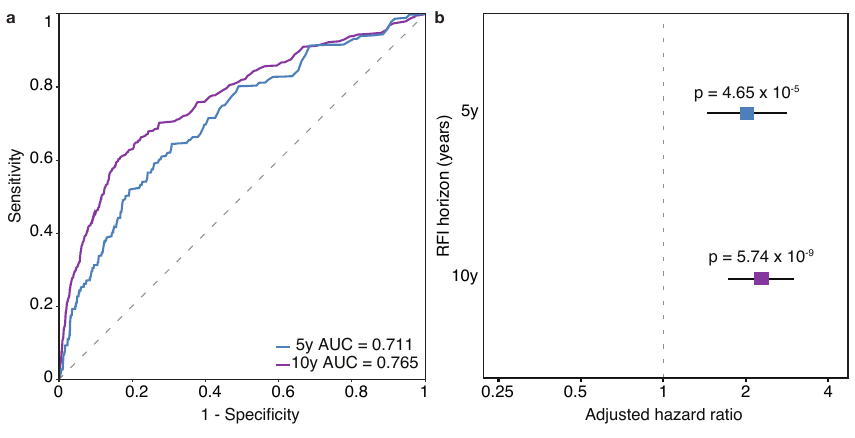}
    \caption{
    \scriptsize
    \textbf{Evaluation of CTX as a prognostic marker.}
    CTX-prognostic predictions for recurrence-free interval (RFI) in the evaluation set (n = 1,994).
    \textbf{a}, Receiver operating characteristic (ROC) curves illustrate the sensitivity and specificity of CTX-prognostic for five-year (blue) and 10-year (red) RFI.
    \textbf{b}, Forest plot shows the adjusted hazard ratio (HR) for CTX-prognostic in a multivariable Cox model adjusting for age at diagnosis, tumor stage, and nodal stage, at the five-year (blue) and 10-year (red) horizons. The center box indicates the HR, and error bars represent 95\% confidence intervals (CIs). p-values are indicated for each predictor in the multivariable models.
    }
    \label{fig:validation_prognostic}
\end{figure}

\clearpage
\begin{figure}[h]
    \centering
    \includegraphics[width=1.0\linewidth]{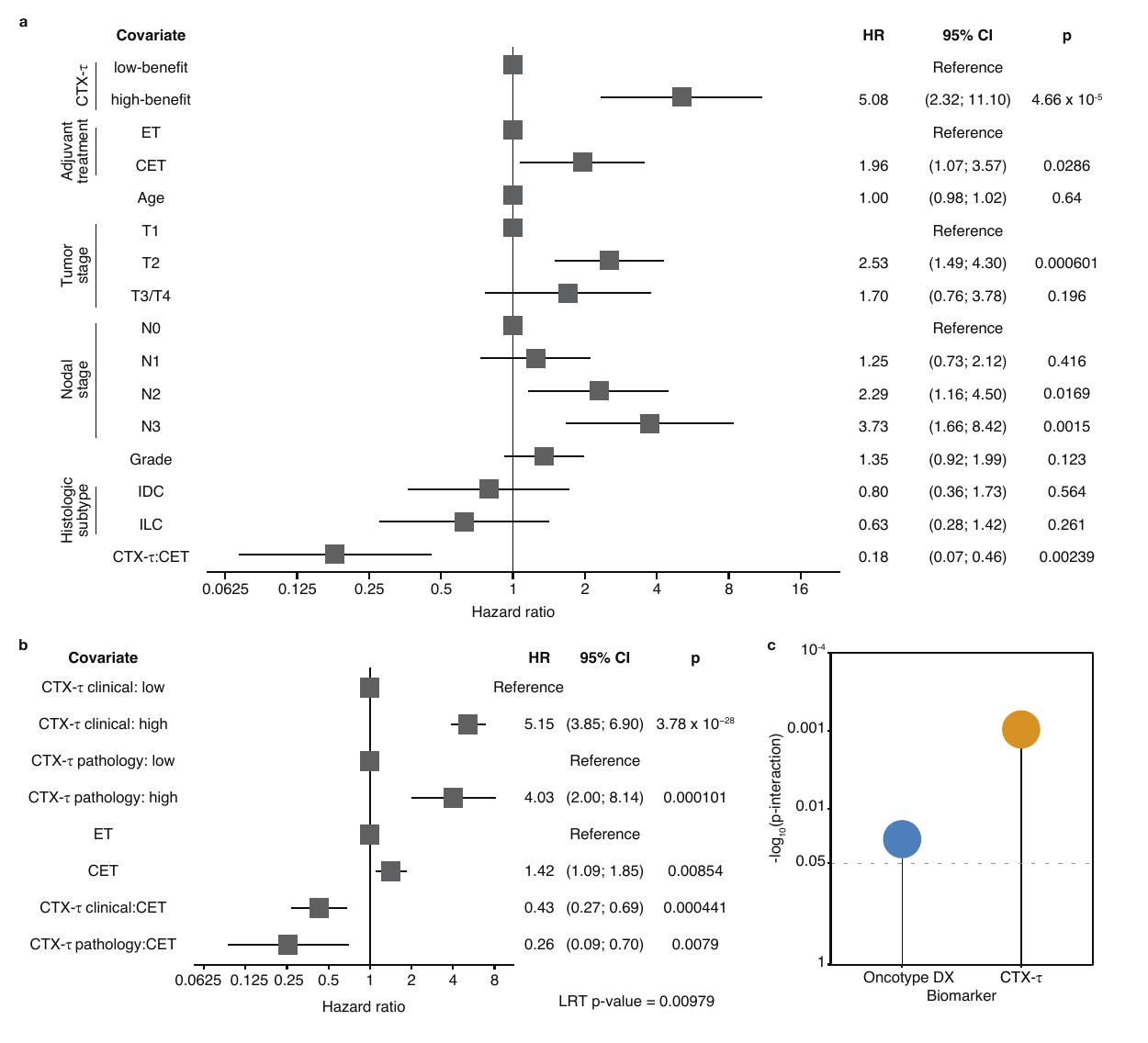}
    \caption{
    \scriptsize
    \textbf{Evaluation of CTX as a predictive marker of chemotherapy benefit.}
    Analyses performed in the evaluation set (n = 1,994) unless otherwise stated. Forest plots show the hazard ratio (HR; center box) with 95\% confidence intervals (CIs; error bars) for each covariate.
    \textbf{a}, Treatment-by-biomarker interaction for chemotherapy and CTX-$\tau$ in an unweighted multivariable Cox model adjusting for CTX-$\tau$ category (low- vs. high-benefit), adjuvant treatment (endocrine therapy [ET] vs. chemoendocrine therapy [CET]), age at diagnosis, tumor stage, nodal stage, grade, and histological subtype (invasive ductal carcinoma [IDC], invasive lobular carcinoma [ILC]), together with the CTX-$\tau$ and chemotherapy interaction term (CTX-$\tau$:CET). The interaction HR quantifies how much the effect of chemotherapy differs between low- and high-benefit patients.
    \textbf{b}, Ablation analysis quantifying the contribution of each input modality to the predictive signal. CTX-$\tau$ scores from the clinical-only and pathology-only models were each dichotomized at the 2\% threshold and entered into a single IPW-adjusted Cox model with chemotherapy treatment status and both treatment-by-biomarker interactions (CTX-$\tau$ clinical:CET, CTX-$\tau$ pathology:CET). The likelihood ratio test (LRT) p-value compares this model against a nested model omitting the pathology interaction, testing whether histopathology adds predictive information beyond clinical features.
    \textbf{c}, Head-to-head comparison of the strength of treatment-effect modification by CTX-$\tau$ and the Oncotype DX recurrence score, in the subset of patients with a recorded Oncotype DX score (n = 983). Bars show the treatment-by-biomarker interaction p-value from an IPW-adjusted Cox model containing the standardized biomarker, chemotherapy treatment status, and their interaction, fit separately for each biomarker in the same patients. The dashed horizontal line indicates p = 0.05.
    }
    \label{fig:supp_benefit}
\end{figure}

\clearpage
\begin{figure}[h]
    \centering
    \includegraphics[width=1.0\linewidth]{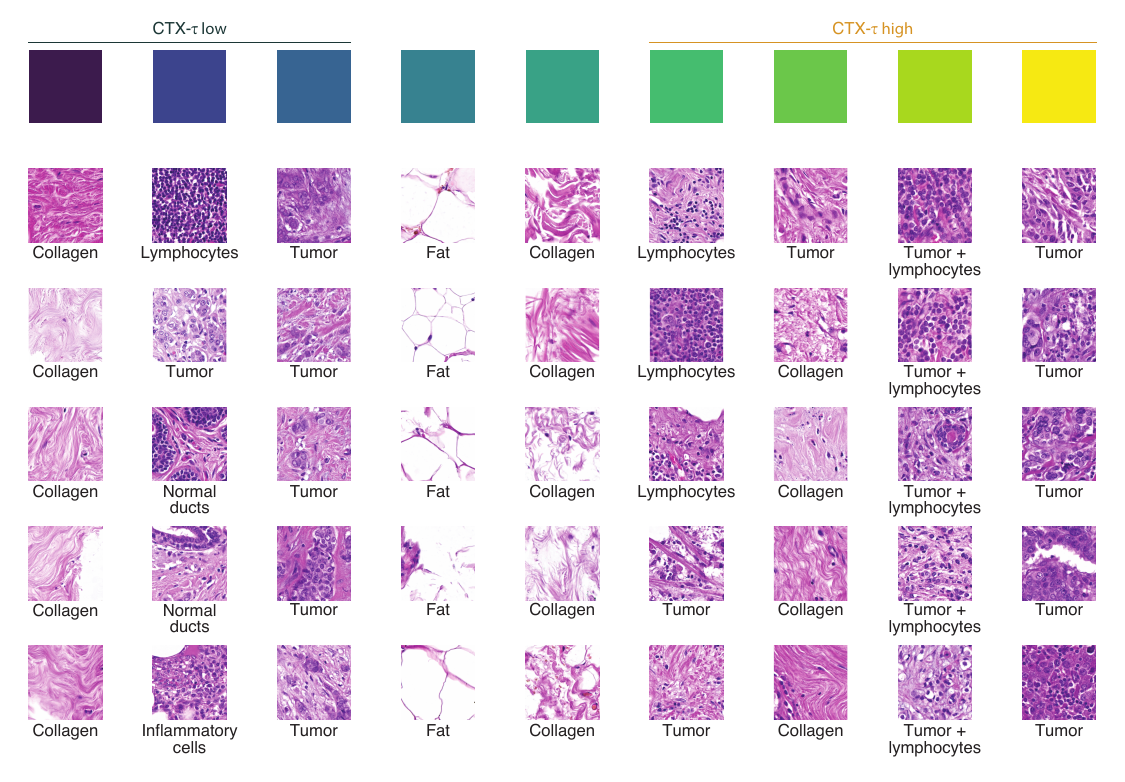}
    \caption{
    \scriptsize
    \textbf{Patch-level annotation of morphological correlates of CTX-$\tau$ predictions.}
    Representative H\&E image patches (224$\times$224 pixels, 0.5\,\textmu m per pixel) from each of the nine $k$-means clusters shown in \textbf{\Cref{fig:explainability}a}, derived from Falcon patch embeddings in the UNIRAD evaluation cohort (n = 362).
    Columns correspond to clusters, ordered left to right by increasing cluster-level CTX-$\tau$ coefficient, from clusters enriched in CTX-$\tau$ low-benefit tumors (dark blue) to those enriched in CTX-$\tau$ high-benefit tumors (bright yellow); the colored square above each column matches the cluster coloring in \textbf{\Cref{fig:explainability}a}.
    Five randomly sampled patches are shown per cluster.
    The label beneath each patch gives the dominant morphological pattern annotated by a board-certified pathologist with subspecialty expertise in breast pathology.
    Clusters associated with low predicted benefit are dominated by fibrotic collagen, adipose tissue, normal ducts, and lymphocytes, with sparse invasive carcinoma; clusters associated with high predicted benefit consist predominantly of nests of invasive carcinoma with high cellularity, alone or admixed with lymphocytes.
    }
    \label{fig:path_explain}
\end{figure}

\clearpage
\begin{figure}[h]
    \centering
    \begin{minipage}[t]{0.495\linewidth}
        \centering
        \includegraphics[width=\linewidth,trim=17mm 0 15mm 0,clip]{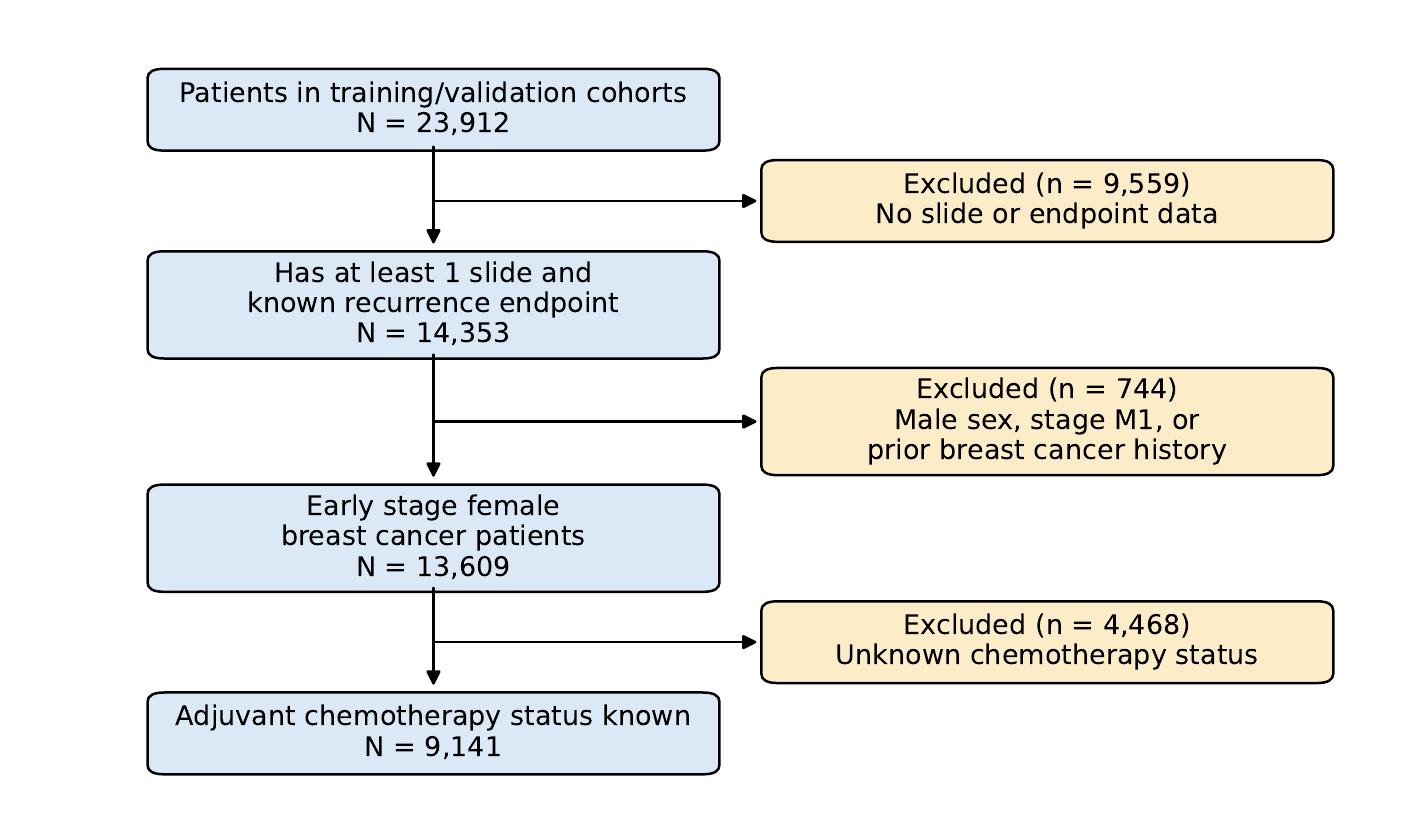}
        \caption*{\scriptsize Development set}
    \end{minipage}
    \hfill
    \begin{minipage}[t]{0.495\linewidth}
        \centering
        \includegraphics[width=\linewidth,trim=17mm 0 15mm 0,clip]{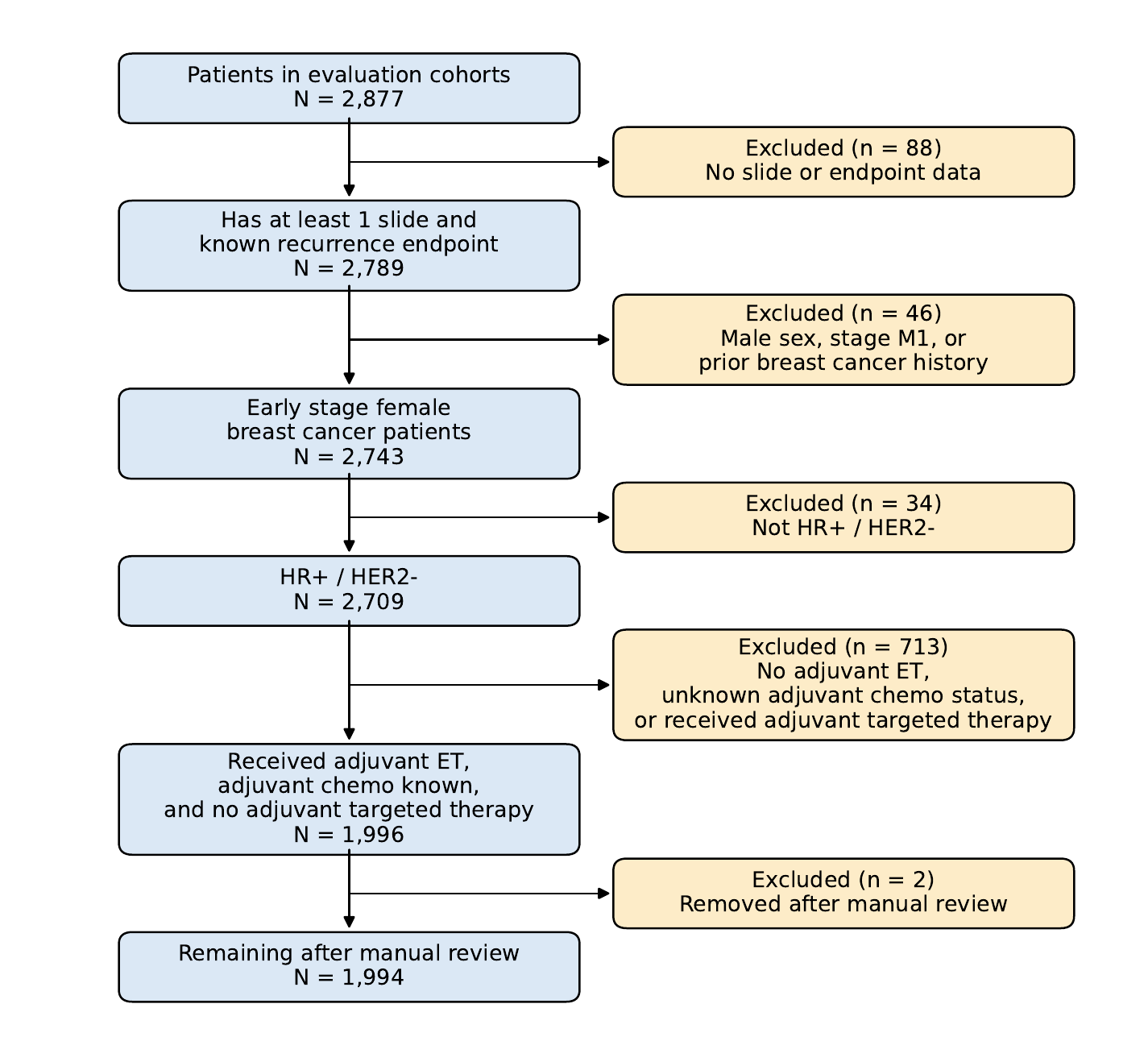}
        \caption*{\scriptsize Evaluation set}
    \end{minipage}
    \caption{
    \scriptsize
    \textbf{Data filtering flow diagrams.}
    Data filtering flow diagrams show the sequential inclusion and exclusion criteria applied to assemble the development set (left, 12 cohorts) and the held-out evaluation set (right, 5 cohorts). Blue boxes indicate the number of patients retained at each step. Orange boxes indicate the number excluded and the reason for exclusion.
    Of 23,912 patients in the development cohorts, 9,141 with at least one whole-slide image, a known recurrence endpoint, early-stage female breast cancer, no prior breast cancer history, and known adjuvant chemotherapy status were retained for model development. Development cohorts were deliberately not restricted by receptor subtype or treatment.
    Of 2,877 patients in the evaluation cohorts, 1,994 were retained after additionally restricting to the intended deployment population --- HR+/HER2$-$ disease, receipt of adjuvant endocrine therapy (ET) with known adjuvant chemotherapy status, no adjuvant targeted therapy --- and excluding slides that failed quality review.
    }
    \label{fig:exclusion_diagrams}
\end{figure}

\setcounter{table}{0}
\renewcommand{\tablename}{Extended Data Table}
\renewcommand{\thetable}{\arabic{table}}

\newpage
\begin{table}[h]
    \centering
    \tiny
    \caption{\textbf{Baseline characteristics of the development and evaluation cohorts.}}
    \label{tab:population}
    \begin{threeparttable}
    \begin{tabular}{@{}lrr@{}}
    \toprule
    \textbf{Characteristic} & \textbf{Development cohort (N=9141)} & \textbf{Evaluation cohort (N=1994)} \\
    \midrule
    \textbf{Age at diagnosis}, median [iqr] & 56.0 [47.0--65.1] & 57.0 [49.0--65.0] \\
    \addlinespace[2pt]
    \textbf{Race} &  &  \\
    \hspace{1em}Asian & 1181 (12.92\%) & 26 (1.30\%) \\
    \hspace{1em}Black or African American & 96 (1.05\%) & 154 (7.72\%) \\
    \hspace{1em}White & 1114 (12.19\%) & 331 (16.60\%) \\
    \hspace{1em}Other/Unknown & 6750 (73.84\%) & 1483 (74.37\%)\\
    \addlinespace[2pt]
    \textbf{ER receptor status} &  &  \\
    \hspace{1em}Negative & 1517 (16.60\%) & 3 (0.15\%) \\
    \hspace{1em}Positive & 6766 (74.02\%) & 1991 (99.85\%) \\
    \hspace{1em}Unknown & 858 (9.39\%) & 0 (0.00\%) \\
    \addlinespace[2pt]
    \textbf{PR receptor status} &  &  \\
    \hspace{1em}Negative & 2360 (25.82\%) & 198 (9.93\%) \\
    \hspace{1em}Positive & 5337 (58.39\%) & 1775 (89.02\%) \\
    \hspace{1em}Unknown & 1444 (15.80\%) & 21 (1.05\%) \\
    \addlinespace[2pt]
    \textbf{HER2 receptor status (by IHC)} &  &  \\
    \hspace{1em}Equivocal (2+) & 157 (1.72\%) & 0 (0.00\%) \\
    \hspace{1em}Negative (0,1+) & 6232 (68.18\%) & 1994 (100.00\%) \\
    \hspace{1em}Positive (3+) & 1228 (13.43\%) & 0 (0.00\%) \\
    \hspace{1em}Unknown & 1524 (16.67\%) & 0 (0.00\%) \\
    \addlinespace[2pt]
    \textbf{T stage} &  &  \\
    \hspace{1em}T0 & 15 (0.16\%) & 0 (0.00\%) \\
    \hspace{1em}T1 & 3604 (39.43\%) & 1189 (59.63\%) \\
    \hspace{1em}T2 & 4119 (45.06\%) & 656 (32.90\%) \\
    \hspace{1em}T3 & 652 (7.13\%) & 112 (5.62\%) \\
    \hspace{1em}T4 & 227 (2.48\%) & 14 (0.70\%) \\
    \hspace{1em}TX & 10 (0.11\%) & 9 (0.45\%) \\
    \hspace{1em}Tis & 371 (4.06\%) & 0 (0.00\%) \\
    \hspace{1em}Unknown & 143 (1.56\%) & 14 (0.70\%) \\
    \addlinespace[2pt]
    \textbf{N stage} &  &  \\
    \hspace{1em}N0 & 4728 (51.72\%) & 1161 (58.22\%) \\
    \hspace{1em}N1 & 2450 (26.80\%) & 638 (32.00\%) \\
    \hspace{1em}N2 & 765 (8.37\%) & 117 (5.87\%) \\
    \hspace{1em}N3 & 498 (5.45\%) & 50 (2.51\%) \\
    \hspace{1em}NX & 89 (0.97\%) & 16 (0.80\%) \\
    \hspace{1em}Unknown & 611 (6.68\%) & 12 (0.60\%) \\
    \addlinespace[2pt]
    \textbf{Ductal histology} &  &  \\
    \hspace{1em}No & 1360 (14.88\%) & 376 (18.86\%) \\
    \hspace{1em}Yes & 6960 (76.14\%) & 1618 (81.14\%) \\
    \hspace{1em}Unknown & 821 (8.98\%) & 0 (0.00\%) \\
    \addlinespace[2pt]
    \textbf{Lobular histology} &  &  \\
    \hspace{1em}No & 7344 (80.34\%) & 1565 (78.49\%) \\
    \hspace{1em}Yes & 976 (10.68\%) & 429 (21.51\%) \\
    \hspace{1em}Unknown & 821 (8.98\%) & 0 (0.00\%) \\
    \addlinespace[2pt]
    \textbf{Recurrence} &  &  \\
    \hspace{1em}No & 7854 (85.92\%) & 1812 (90.87\%) \\
    \hspace{1em}Yes & 1287 (14.08\%) & 182 (9.13\%) \\
    \addlinespace[2pt]
    \textbf{Distant recurrence} &  &  \\
    \hspace{1em}No & 8323 (91.05\%) & 1851 (92.83\%) \\
    \hspace{1em}Yes & 818 (8.95\%) & 143 (7.17\%) \\
    \addlinespace[2pt]
    \textbf{Death} &  &  \\
    \hspace{1em}No & 8303 (90.83\%) & 1803 (90.42\%) \\
    \hspace{1em}Yes & 734 (8.03\%) & 191 (9.58\%) \\
    \hspace{1em}Unknown & 104 (1.14\%) & 0 (0.00\%) \\
    \addlinespace[2pt]
    \textbf{Received adjuvant endocrine therapy} &  &  \\
    \hspace{1em}No & 2374 (25.97\%) & 0 (0.00\%) \\
    \hspace{1em}Unknown & 237 (2.59\%) & 0 (0.00\%) \\
    \hspace{1em}Yes & 6530 (71.44\%) & 1994 (100.00\%) \\
    \addlinespace[2pt]
    \textbf{Received adjuvant chemotherapy} &  &  \\
    \hspace{1em}No & 4812 (52.64\%) & 1183 (59.33\%) \\
    \hspace{1em}Yes & 4329 (47.36\%) & 811 (40.67\%) \\
    \addlinespace[2pt]
    \textbf{Follow-up time}, median [iqr] & 4.3 [2.0--6.8] & 6.3 [4.3--10.3] \\
    \bottomrule
    \end{tabular}
    \begin{tablenotes}[flushleft]\footnotesize
    \item IQR, interquartile range; ER, estrogen receptor; PR, progesterone receptor; HER2, human epidermal growth factor receptor 2; IHC, immunohistochemistry.
    \end{tablenotes}
    \end{threeparttable}
\end{table}

\newpage
\begin{table}[h]
    \centering
    \tiny
    \caption{\textbf{Baseline characteristics of the evaluation cohorts.}}
    \label{tab:test}
    \makebox[\textwidth][c]{%
    \begin{threeparttable}
    \begin{tabular}{@{}lrrrrr@{}}
    \toprule
    \textbf{Characteristic} & \textbf{Karmanos cohort (N=147)} & \textbf{UChicago cohort (N=435)} & \textbf{Dordrecht cohort (N=582)} & \textbf{Roswell Park cohort (N=468)} & \textbf{UNIRAD cohort (N=362)} \\
    \midrule
    \textbf{Age at diagnosis}, median [iqr] & 59.0 [50.0--66.0] & 56.0 [48.0--65.0] & 60.0 [51.0--68.0] & 57.0 [50.0--64.0] & 52.0 [46.0--62.0] \\
    \addlinespace[2pt]
    \textbf{Race} &  &  &  &  &  \\
    \hspace{1em}Black or African American & 45 (30.61\%) & 109 (25.06\%) & 0 (0.00\%) & 0 (0.00\%) & 0 (0.00\%) \\
    \hspace{1em}White & 41 (27.89\%) & 290 (66.67\%) & 0 (0.00\%) & 0 (0.00\%) & 0 (0.00\%) \\
    \hspace{1em}Asian & 0 (0.00\%) & 26 (5.98\%) & 0 (0.00\%) & 0 (0.00\%) & 0 (0.00\%) \\
    \hspace{1em}Other/Unknown & 61 (41.50\%) & 10 (2.30\%) & 582 (100.00\%) & 468 (100.00\%) & 362 (100.00\%) \\
    \addlinespace[2pt]
    \textbf{ER receptor status} &  &  &  &  &  \\
    \hspace{1em}Negative & 1 (0.68\%) & 1 (0.23\%) & 0 (0.00\%) & 1 (0.21\%) & 0 (0.00\%) \\
    \hspace{1em}Positive & 146 (99.32\%) & 434 (99.77\%) & 582 (100.00\%) & 467 (99.79\%) & 362 (100.00\%) \\
    \addlinespace[2pt]
    \textbf{PR receptor status} &  &  &  &  &  \\
    \hspace{1em}Negative & 8 (5.44\%) & 46 (10.57\%) & 45 (7.73\%) & 43 (9.19\%) & 56 (15.47\%) \\
    \hspace{1em}Positive & 139 (94.56\%) & 389 (89.43\%) & 516 (88.66\%) & 425 (90.81\%) & 306 (84.53\%) \\
    \hspace{1em}Unknown & 0 (0.00\%) & 0 (0.00\%) & 21 (3.61\%) & 0 (0.00\%) & 0 (0.00\%) \\
    \addlinespace[2pt]
    \textbf{HER2 receptor status (by IHC)} &  &  &  &  &  \\
    \hspace{1em}Negative (0,1+) & 147 (100.00\%) & 435 (100.00\%) & 582 (100.00\%) & 468 (100.00\%) & 362 (100.00\%) \\
    \addlinespace[2pt]
    \textbf{T stage} &  &  &  &  &  \\
    \hspace{1em}T1 & 83 (56.46\%) & 283 (65.06\%) & 351 (60.31\%) & 361 (77.14\%) & 111 (30.66\%) \\
    \hspace{1em}T2 & 51 (34.69\%) & 123 (28.28\%) & 213 (36.60\%) & 97 (20.73\%) & 172 (47.51\%) \\
    \hspace{1em}T3 & 3 (2.04\%) & 16 (3.68\%) & 14 (2.41\%) & 10 (2.14\%) & 69 (19.06\%) \\
    \hspace{1em}T4 & 1 (0.68\%) & 4 (0.92\%) & 4 (0.69\%) & 0 (0.00\%) & 5 (1.38\%) \\
    \hspace{1em}Unknown & 9 (6.12\%) & 9 (2.07\%) & 0 (0.00\%) & 0 (0.00\%) & 5 (1.38\%) \\
    \addlinespace[2pt]
    \textbf{N stage} &  &  &  &  &  \\
    \hspace{1em}N0 & 98 (66.67\%) & 316 (72.64\%) & 347 (59.62\%) & 396 (84.62\%) & 4 (1.10\%) \\
    \hspace{1em}N1 & 34 (23.13\%) & 105 (24.14\%) & 231 (39.69\%) & 68 (14.53\%) & 200 (55.25\%) \\
    \hspace{1em}N2 & 2 (1.36\%) & 3 (0.69\%) & 0 (0.00\%) & 1 (0.21\%) & 111 (30.66\%) \\
    \hspace{1em}N3 & 2 (1.36\%) & 0 (0.00\%) & 0 (0.00\%) & 1 (0.21\%) & 47 (12.98\%) \\
    \hspace{1em}Unknown & 2 (1.36\%) & 1 (0.23\%) & 4 (0.69\%) & 2 (0.43\%) & 0 (0.00\%) \\
    \addlinespace[2pt]
    \textbf{Ductal histology} &  &  &  &  &  \\
    \hspace{1em}No & 19 (12.93\%) & 83 (19.08\%) & 108 (18.56\%) & 82 (17.52\%) & 84 (23.20\%) \\
    \hspace{1em}Yes & 128 (87.07\%) & 352 (80.92\%) & 474 (81.44\%) & 386 (82.48\%) & 278 (76.80\%) \\
    \addlinespace[2pt]
    \textbf{Lobular histology} &  &  &  &  &  \\
    \hspace{1em}No & 128 (87.07\%) & 324 (74.48\%) & 483 (82.99\%) & 356 (76.07\%) & 274 (75.69\%) \\
    \hspace{1em}Yes & 19 (12.93\%) & 111 (25.52\%) & 99 (17.01\%) & 112 (23.93\%) & 88 (24.31\%) \\
    \addlinespace[2pt]
    \textbf{Recurrence} &  &  &  &  &  \\
    \hspace{1em}No & 132 (89.80\%) & 410 (94.25\%) & 542 (93.13\%) & 422 (90.17\%) & 306 (84.53\%) \\
    \hspace{1em}Yes & 15 (10.20\%) & 25 (5.75\%) & 40 (6.87\%) & 46 (9.83\%) & 56 (15.47\%) \\
    \addlinespace[2pt]
    \textbf{Distant recurrence} &  &  &  &  &  \\
    \hspace{1em}No & 137 (93.20\%) & 418 (96.09\%) & 550 (94.50\%) & 435 (92.95\%) & 311 (85.91\%) \\
    \hspace{1em}Yes & 10 (6.80\%) & 17 (3.91\%) & 32 (5.50\%) & 33 (7.05\%) & 51 (14.09\%) \\
    \addlinespace[2pt]
    \textbf{Received adjuvant endocrine therapy} &  &  &  &  &  \\
    \hspace{1em}Yes & 147 (100.00\%) & 435 (100.00\%) & 582 (100.00\%) & 468 (100.00\%) & 362 (100.00\%) \\
    \addlinespace[2pt]
    \textbf{Received adjuvant chemotherapy} &  &  &  &  &  \\
    \hspace{1em}No & 108 (73.47\%) & 322 (74.02\%) & 329 (56.53\%) & 352 (75.21\%) & 72 (19.89\%) \\
    \hspace{1em}Yes & 39 (26.53\%) & 113 (25.98\%) & 253 (43.47\%) & 116 (24.79\%) & 290 (80.11\%) \\
    \addlinespace[2pt]
    \textbf{Received adjuvant targeted therapy} &  &  &  &  &  \\
    \hspace{1em}No & 147 (100.00\%) & 435 (100.00\%) & 582 (100.00\%) & 468 (100.00\%) & 362 (100.00\%) \\
    \addlinespace[2pt]
    \textbf{Death} &  &  &  &  &  \\
    \hspace{1em}No & 145 (98.64\%) & 428 (98.39\%) & 516 (88.66\%) & 376 (80.34\%) & 338 (93.37\%) \\
    \hspace{1em}Yes & 2 (1.36\%) & 7 (1.61\%) & 66 (11.34\%) & 92 (19.66\%) & 24 (6.63\%) \\
    \addlinespace[2pt]
    \textbf{Follow-up time}, median [iqr] & 4.9 [4.1--6.2] & 7.0 [4.5--10.7] & 5.9 [4.3--8.2] & 13.2 [10.4--14.9] & 4.6 [3.2--5.9] \\
    \bottomrule
    \end{tabular}
    \begin{tablenotes}[flushleft]\footnotesize
    \item IQR, interquartile range; ER, estrogen receptor; PR, progesterone receptor; HER2, human epidermal growth factor receptor 2; IHC, immunohistochemistry.
    \end{tablenotes}
    \end{threeparttable}
    }
\end{table}

\newpage
\begin{table}[h]
    \centering
    \caption{{\bf CTX hyperparameter search space.} Continuous parameters were sampled from Uniform or Log-uniform distributions over the specified ranges; discrete sets were sampled uniformly.}
    \label{tab:hyperparameters}
    \begin{tabular}{ll}
    \toprule
    \textbf{Hyperparameter} & \textbf{Search space} \\
    \midrule
    \multicolumn{2}{l}{\textit{Optimization}} \\
    Modality branch weights ($\lambda_1, \lambda_2$) & Uniform (0.0, 1.0) \\
    Elastic-net mixing weight ($\gamma$) & Uniform (0.0, 1.0) \\
    Elastic-net regularization weight ($\alpha$) & Log-uniform ($10^{-6}$, $10^{-3}$) \\
    Learning rate & Log-uniform ($10^{-3}$, $3 \times 10^{-2}$) \\
    \addlinespace
    \multicolumn{2}{l}{\textit{Balancing penalty (MMD)}} \\
    MMD divergence weight target ($\alpha_{\mathrm{MMD}}$) & Uniform (0.0, 0.99) \\
    \addlinespace
    \multicolumn{2}{l}{\textit{Survival model}} \\
    Number of time cut points ($K$) & Uniform integer (4, 8) \\
    \addlinespace
    \multicolumn{2}{l}{\textit{Tabular ResNet (clinical branch)}} \\
    Number of residual blocks ($n_{\text{blocks}}$) & Uniform integer (1, 6) \\
    Block width ($d_{\text{block}}$) & Log-uniform integer (4, 32) \\
    Hidden dimension multiplier ($\gamma_{\text{hidden\_mult}}$) & Uniform (1.0, 3.0) \\
    Clinical dropout rate & Uniform (0.0, 0.3) \\
    \addlinespace
    \multicolumn{2}{l}{\textit{Multiple instance learning (pathology branch)}} \\
    Gated-attention hidden dimension ($H$) & $\{1,2,3,4\}$ \\
    Pathology dropout rate & Uniform (0.0, 0.3) \\
    \addlinespace
    \multicolumn{2}{l}{\textit{Shared between clinical and pathology branches}} \\
    Per-branch projection / representation ($d_\text{int}$) & Log-uniform integer (4, 512) \\
    Output-head FC hidden-dim multiplier ($\gamma_\text{fc}$) & Uniform (1.0, 3.0) \\
    Output-head dropout rate & Uniform (0.0, 0.3) \\
    \addlinespace
    \multicolumn{2}{l}{\textit{Multi-modal fusion}} \\
    Fusion attention hidden-dim multiplier ($\gamma_{\text{att}}$) & Uniform (0.1, 3.0) \\
    \bottomrule
    \end{tabular}
\end{table}

\clearpage
\newpage

\appendix

\part{Appendix}
\section{Supplementary Methods} \label{sec:supplementary_methods}

\setcounter{figure}{0}
\renewcommand{\figurename}{Supplemental Fig.}
\renewcommand{\thefigure}{\arabic{figure}}
\setcounter{table}{0}
\renewcommand{\tablename}{Supplemental Table}
\renewcommand{\thetable}{\arabic{table}}

This section is split into four parts. First, we describe the data in \textbf{\cref{sec:methods:data}}. Second, we define notation and the two quantities CTX outputs in \textbf{\cref{sec:methods:notation}}. Third, we review our modeling choices in detail in \textbf{\cref{sec:methods:ctx}}. Finally, we describe how CTX is evaluated in \textbf{\cref{sec:methods:evaluation}}.

\subsection{Datasets}  \label{sec:methods:data}

To develop and evaluate the model, we assembled a retrospectively collected dataset containing 11,135 patients from 17 independent cohorts. 
We restricted to female patients with early-stage breast cancer who had known adjuvant chemotherapy status, known RFI outcomes, no prior history of breast cancer, and at least one available pathology slide. 
Twelve cohorts (n = 9,141) were used for model development --- the {\it development set}. The other five cohorts (n = 1,994) were held out for evaluation --- the {\it evaluation set}. 
In the evaluation set, we further restricted the analysis to HR+/HER2$-$ patients who received adjuvant endocrine therapy (ET) or chemoendocrine therapy (CET).
Further details of the exclusion criteria are visualized in Extended Data~\Cref{fig:exclusion_diagrams}. 
A detailed breakdown of patient characteristics can be found in Extended Data~\Cref{tab:population}. 
All of our datasets are observational cohorts besides UNIRAD, which is a phase III clinical study with randomized assignment of the targeted therapy everolimus, but non-random assignment of chemotherapy~\citep{bachelot2022everolimus}.
The Oncotype DX recurrence score was available for 983 patients across the evaluation set (\Cref{tab:odx_mmp_categories}).

\begin{table}[h]
    \centering
    \small
    \caption{\textbf{Oncotype DX score category distributions across the evaluation datasets.}}
    \label{tab:odx_mmp_categories}
    \begin{threeparttable}
    \begin{tabular}{@{}lrrrrr@{}}
    \toprule
    & \textbf{Dordrecht} & \textbf{Roswell Park} & \textbf{UChicago} & \textbf{UNIRAD} & \textbf{Karmanos} \\
    \midrule
    \hspace{1em}Low & 0 (0\%) & 102 (22\%) & 73 (17\%) & 0 (0\%) & 33 (22\%) \\
    \hspace{1em}Intermediate & 0 (0\%) & 297 (63\%) & 231 (53\%) & 0 (0\%) & 85 (58\%) \\
    \hspace{1em}High & 0 (0\%) & 69 (15\%) & 65 (15\%) & 0 (0\%) & 28 (19\%) \\
    \hspace{1em}Not tested & 582 (100\%) & 0 (0\%) & 66 (15\%) & 362 (100\%) & 1 (1\%) \\
    \bottomrule
    \end{tabular}
    \begin{tablenotes}[flushleft]\footnotesize
    \item RS, recurrence score; ODX, Oncotype DX; ODX categories use the clinical cutoffs: Low (RS 0--10), Intermediate (RS 11--25), High (RS $\ge$ 26).
    \end{tablenotes}
    \end{threeparttable}
\end{table}

\subsection{Notation and CTX outputs} \label{sec:methods:notation}

Here we define mathematical notation used throughout. Capitalized letters denote random variables, boldface denote vectors, hats ($\hat f$) denote estimates, and a subscript $i$ denotes patients.

For each patient, we observe data that we represent mathematically as ($\bm X, A, \widetilde T, \Delta$), where \(\bm X\in\mathbb{R}^d\) denotes a vector of clinicopathological and morphological features and $A\in\{0,1\}$ denotes treatment. Here, $T\in\mathbb{R}_{+}$ denotes the time between diagnosis and recurrence event, and $C \in \mathbb{R}_{+}$ denotes censoring time, from which we only observe the follow-up time $\widetilde T = \min(T, C)$ and the event indicator $\Delta = \mathbf{1}(T \leq C)$. CET corresponds to $A=1$ and ET to $A=0$. Our primary outcome of interest is the five-year recurrence-free interval (RFI), that is, $Y\coloneqq\mathbf{1}(T\geq5)$. 

We denote the five-year RFI probability for a patient with features $\bm x$ and treatment $a$ by
\begin{equation*} 
    s^{(a)} (\bm x) \coloneqq \mathbb{P}(T\geq5 \mid A=a, \bm X =\bm x)=\mathbb{E}( Y \mid A=a, \bm X = \bm x ). 
\end{equation*}
Our estimation target is $\tau(\bm x)$, the difference between RFI probabilities under CET and ET. That is,
\begin{align*}
    \tau(\bm x) \coloneqq s^{(1)}(\bm x) - s^{(0)}(\bm x).
\end{align*}
We also refer to this quantity --- that CTX is built to estimate --- as personalized chemotherapy benefit. Specifically, for a patient with features $\bm x$, CTX outputs
\begin{equation*}
    \hat \tau(\bm x) = \hat s^{(1)}(\bm x) - \hat s^{(0)}(\bm x). 
\end{equation*}
This estimate is the CTX-$\tau$ score (\Cref{subsec:ctx}).

For each patient, only the outcome under the treatment that the patient actually received is observed. We refer to the model prediction corresponding to this observed treatment assignment as the \emph{factual prediction}. For a patient with features $\bm x$ and observed treatment $a$, the factual prediction is
\[
\hat s^{(a)}(\bm x) :=
\begin{cases}
  \hat s^{(1)}(\bm x), & a = 1,\\[2pt]
  \hat s^{(0)}(\bm x), & a = 0.
\end{cases}
\]
Thus, for patients who received CET, the factual prediction is $\hat{s}^{(1)}(\bm x)$, whereas for patients who received ET, it is $\hat{s}^{(0)}(\bm x)$. Factual predictions can be compared directly against observed outcomes and are therefore used to evaluate prognostic discrimination and calibration. We define CTX-prognostic as the factual recurrence risk: 
\[
\hat r^{(a)}(\bm x) \;\equiv\; 1 - \hat s^{(a)}(\bm x) :=
\begin{cases}
  1 - \hat s^{(1)}(\bm x), & a = 1,\\[2pt]
  1 - \hat s^{(0)}(\bm x), & a = 0.
\end{cases}
\]
Finally, we define CTX-risk, the baseline risk, as the model-predicted recurrence risk under ET: 
\[
\hat r^{(0)}(\bm x) \coloneqq 1 - \hat s^{(0)}(\bm x).
\]

\subsection{Model development} \label{sec:methods:ctx}

Building the CTX model required addressing four methodological challenges inherent to estimating personalized chemotherapy benefit from observational data. \textbf{Supplemental \Cref{tab:challenges}} summarizes each challenge and the component of our approach that addresses it. In this section we describe each of these components in detail.

\begin{table}[htbp]
\centering
\small
\caption{\textbf{Methodological challenges and solutions.} H\&E, hematoxylin and eosin; WSI, whole slide image.}
\label{tab:challenges}
\begin{tabular}{@{}c >{\raggedright\arraybackslash}p{0.44\linewidth} >{\raggedright\arraybackslash}p{0.44\linewidth}@{}}
\toprule
& \textbf{Challenge} & \textbf{Our solution} \\
\midrule
1 & Only unlabelled H\&E WSIs are commonly available, so robust and generalizable representations must be learned without supervision & A multi-modal architecture built on a self-supervised pathology foundation model (\textbf{\cref{sec:methods:multimodal_architecture,sec:methods:falcon}}) \\
\addlinespace
2 & Treatment assignment is not randomized in observational data, so patients who received chemotherapy differ systematically from those who did not & Treatment-specific output heads and a between-arm balancing penalty that improve extrapolation across treatment groups (\textbf{\cref{sec:methods:multimodal_architecture,sec:methods:objective}}) \\
\addlinespace
3 & The outcome is a censored, time-to-event endpoint & A tailored training objective (\textbf{\cref{sec:methods:objective}}) \\
\addlinespace
4 & The data are pooled across heterogeneous cohorts & Multi-source cross-validation (\textbf{\cref{sec:methods:mscv}}) \\
\bottomrule
\end{tabular}
\end{table}

\subsubsection{Multi-modal model architecture}\label{sec:methods:multimodal_architecture}

The multi-modal architecture of our model consists of three branches: a \textit{clinical branch}, which processes clinical features into a clinical representation; a \textit{pathology branch}, which processes digitized hematoxylin and eosin H\&E WSIs into a pathology representation; and a \textit{multi-modal branch}, which processes a fusion of the clinical and pathology representations. Each branch terminates in its own treatment-specific output heads. The chemotherapy benefit estimate $\hat \tau$ is derived from the output heads of the multi-modal branch (see \textbf{Supplemental~\Cref{fig:multimodal-architecture}} and \textbf{\cref{sec:methods:causal}}), while the clinical and pathology output heads provide auxiliary predictions used only for training (described below).

\begin{figure}[H]
    \centering
    \includegraphics[width=\textwidth]{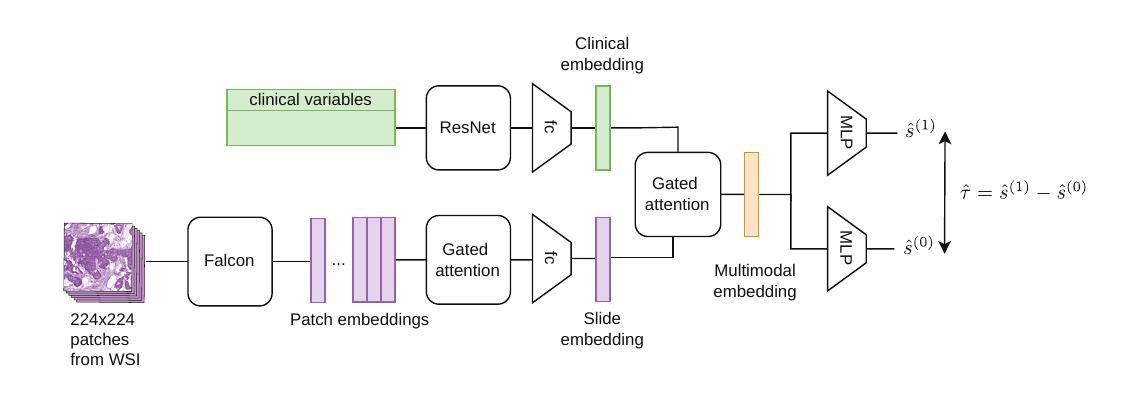}
    \caption{\textbf{Architecture of the proposed Multi-modal model for estimating chemotherapy benefit.}
    Our model takes two input modalities for each patient: clinical variables and histopathology image. The clinical variables are encoded into a clinical embedding using a tabular ResNet, while patches of the image are converted into patch embeddings and aggregated into a slide-level embedding through gated attention. Then the clinical and slide representations are fused through gated attention, and passed to output heads estimating recurrence-free probability under chemoendocrine therapy $\hat{s}^{(1)}$ and endocrine therapy alone $\hat{s}^{(0)}$. The predicted chemotherapy benefit ($\hat{\tau}$) is the difference $\hat s^{(1)} - \hat s^{(0)}$. Auxiliary output heads of the clinical and pathology branches are omitted for clarity. FC denotes a single fully connected (linear) layer, whereas the MLP consists of two fully connected layers with a ReLU activation between them.
}
    \label{fig:multimodal-architecture}
\end{figure}

\paragraph{Clinical.}

For the clinical branch, input features were: tumor stage (T), nodal stage (N), patient age, AI genomic risk, ER/PR/HER2 staining statuses, and ductal and lobular histology. 
Age and AI genomic risk (\textbf{\cref{sec:research_odx}}) are real-valued features, and the rest are categorical. 
Missing age values were imputed using the cohort mean, while unknown categorical values were treated as separate categories. 
Categorical variables were one-hot encoded and concatenated with real-valued features into a single vector denoted by $\mathbf{x}_{\text{clinical}}$. 
Once concatenated, the clinical branch of the model then uses a tabular ResNet~\cite{gorishniy2021revisiting} to encode $\mathbf{x}_{\text{clinical}}$ into an embedding $\mathbf{z}_{\text{clinical}}$. 
The final clinical embedding $\mathbf{z}_\textrm{clinical}$ is used as input to the multi-modal fusion component as well as a clinical-only output heads.

\paragraph{Pathology.} Each WSI is first divided into patches, which are next transformed into a collection of high-dimensional embedding vectors using our pretrained pathology foundation model (described in \textbf{\cref{sec:methods:falcon}}). Then all patch embeddings are aggregated into a single WSI-level representation using a gated attention {\it multiple instance learning} (MIL) layer~\citep{pmlr-v80-ilse18a}. 

As the clinical representation, the pathology representation, $\mathbf{z}_{\mathrm{pathology}}$, is subsequently passed into the multi-modal fusion layer, as well as into a pathology-only output heads.

\paragraph{Multi-modal.} The clinical and pathology embeddings, $\mathbf{z}_{\mathrm{clinical}}$ and $\mathbf{z}_{\mathrm{pathology}}$, are fused together using a gated attention module to produce a multi-modal embedding, $\mathbf{z}_{\mathrm{multimodal}}$.

\paragraph{Treatment-specific output heads with discrete-time hazard parameterization.}
The output heads of our model are designed to predict the likelihood of recurrence events.
Extending the shared-representation, treatment-specific-heads design of~\citet{shalit_icml_2017} to the multi-modal setting, we pair each modality $m \in \{\mathrm{clinical}, \mathrm{pathology}, \mathrm{multimodal}\}$ and each treatment $a \in \{0,1\}$ to a separate output head $g^{(m, a)}$.
The output head $g^{(m, a)}$ is a small feedforward network that takes in the embedding vector $\mathbf{z}_{m}$ and outputs discrete-time logistic hazards. 
The multi-modal heads produce CTX's outputs; the clinical-only and pathology-only heads provide auxiliary outputs used only for training (\textbf{\cref{sec:methods:objective}}).

Hazards are defined by partitioning time as follows. 
We place $K$ cut points $c_0 < c_1 < \cdots < c_{K-1}$ on the time axis, creating $K-1$ time bins. 
The starting cut point is $c_0 = 0$ and the final cut point, $c_{K-1}$, is the largest observed follow-up time in the development set. 
The interior cuts ($c_1, \dots, c_{K-2}$) are placed at evenly spaced quantiles of the event durations in the development set, and then rescaled piecewise-linearly (separately below and above the five-year mark) so that exactly one interior cut coincides with $t = 5$ years.

Each output head $g^{(m,a)}$ produces a vector of $K$ logits, $\bm \phi^{(m,a)}\in\mathbb{R}^K$. The first logit, anchored at the boundary cut $c_0 = 0$, is discarded at prediction time, where the boundary condition $S(0) = 1$ is enforced directly. The remaining $K - 1$ logits parameterize the conditional hazards on the $K - 1$ intervals $(c_{k-1}, c_k]$ for $k = 1, \ldots, K-1$ (note that $K$ is a hyperparameter, see \Cref{tab:hyperparameters}). These logits are mapped to survival hazards by the sigmoid transformation, and the recurrence-free probability at time bin $k$ is obtained by accumulating the hazards: 
\begin{align*}
    h_j^{(m,a)} &= \sigma(\bm \phi_{j+1}^{(m,a)}), \\
    S_k^{(m,a)} &= \prod_{j=1}^{k} (1 - h_j^{(m,a)}).
\end{align*}
The five-year recurrence-free probability is obtained by choosing the time bin $k$ whose right-boundary is exactly at 5 years (this bin must exist, by definition of our interior cut scheme). Our final treatment-specific predictions are 
$$\hat{s}^{(1)}(\bm x)=S_{k_5}^{(\mathrm{multimodal},1)}\quad \text{and}\quad \hat{s}^{(0)}(\bm x)=S_{k_5}^{(\mathrm{multimodal},0)},$$
where $k_5$ is the index of the five-year time bin. Their difference is the CTX-$\tau$ score $\hat \tau$ (\textbf{\cref{sec:methods:notation}}).

\subsubsection{Training objective} \label{sec:methods:objective}

The training objective for CTX is carefully designed to address two key difficulties.
The first difficulty is right-censoring in time-to-event labels. The second is distributional discrepancies between treatment groups. 
Each difficulty is addressed with a specific loss term. 
These terms are then combined in a single loss function and optimized using a gradient-based optimization strategy.

\paragraph{Per-treatment survival loss.}
For a patient $i$ with event indicator $\delta_i \in \{0, 1\}$, observed treatment $a_i$, and $d_i$ the index of the time bin containing the observed follow-up time $\widetilde t_i$, consider the target labels $y_{i, k} = \mathbf{1}[k = d_i]\, \delta_i$ where $k=1,\dots,K-1$ is the bin index. 
The negative log-likelihood (NLL) for patient $i$ under modality $m$, evaluated with respect to the factual treatment head, is
\begin{align}
\mathcal{L}^{(m)}_{\mathrm{NLL}, i}
\;=\;
\sum_{k=1}^{d_i}
\mathrm{BCE}\!\left(h_k^{(m,a_i)},\; y_{i, k}\right),
\end{align}
where $\mathrm{BCE}$ denotes binary cross-entropy. Summing these terms over $k \le d_i$ accounts for right-censoring: intervals after a patient's last observed follow-up contribute nothing to the loss.
For censored patients, summing to $d_i$ treats the censoring time as occurring at the right boundary of that bin.

The outcome loss aggregates the per-modality factual NLLs over the $n$ patients in the development set as a weighted mean,
\begin{align}
\mathcal{L}_{\mathrm{out}}
\;=\;
\frac{1}{n} \sum_{i=1}^{n} \sum_m
\lambda_m \mathcal{L}_{\mathrm{NLL}, i}^{(m)},
\end{align}
where the modality weights are reparameterized based on two scalars $\lambda_1, \lambda_2 \in [0, 1]$ as
\begin{align}
\lambda_{\mathrm{multimodal}} = \lambda_1, \quad
\lambda_{\mathrm{clinical}} = (1 - \lambda_1)\lambda_2, \quad
\lambda_{\mathrm{pathology}} = (1 - \lambda_1)(1 - \lambda_2),
\end{align}
guaranteeing $\sum_m \lambda_m = 1$. Both $\lambda_1$ and $\lambda_2$ are hyperparameters (see \Cref{tab:hyperparameters}).

\paragraph{Maximum mean discrepancy balancing penalty.}
The maximum mean discrepancy (MMD) is an integral probability metric used to quantify the difference between two probability distributions~\citep{muller1997integral}.
In our case, we are interested in the difference between the distributions of multi-modal representations of CET patients and ET patients.
The MMD between the representation distributions of the two treatment groups is a squared distance of kernel mean embeddings in the reproducing kernel Hilbert space $\mathcal{H}$ induced by a kernel $\kappa$~\citep{gretton2012kernel}. That is,
\begin{align} \label{eq:mmd}
\mathcal{L}_{\mathrm{MMD}}
\;&=\;
\lVert \bm{\mu}_\mathcal{D} - \bm{\mu}_{\mathcal{D}'} \rVert_\mathcal{H}^2, \\
\bm{\mu}_{\mathcal{D}}  &= \frac{1}{|\mathcal{D}|}\sum_{i\in\mathcal{D}}\psi(\bm z_i),
   &\mathcal{D}  = \{i : a_i = 0\},\\
\bm{\mu}_{\mathcal{D}'} &= \frac{1}{|\mathcal{D}'|}\sum_{i\in\mathcal{D}'}\psi(\bm z_i),
 &\mathcal{D}' = \{i : a_i = 1\},
\end{align}
where $\kappa(\bm z,\bm z') = \langle \psi(\bm z), \psi(\bm z')\rangle_{\mathcal{H}}$ is the
kernel, $\psi$ is its associated feature map, and $\bm z_i$ is the multi-modal
representation vector (previously denoted $\bm z_\mathrm{multimodal}$)
corresponding to patient $i$. Minimizing $\mathcal{L}_{\mathrm{MMD}}$ reduces covariate shift (in the embedding space) between the two treatment arms, encouraging a representation under which outcomes can be predicted accurately for either arm~\citep{shalit_icml_2017}.

\paragraph{Overall training objective.}
The overall training loss $\mathcal{L}$ combines the outcome and balancing terms with elastic-net regularization on the model parameters $\bm \theta$, which comprise the trainable weights of CTX's clinical and pathology branches, fusion module, and output heads:
\begin{equation}
\begin{aligned}
\mathcal{L}
&=
(1 - \alpha)\Big[(1 - \alpha_{\mathrm{MMD}})\,\mathcal{L}_{\mathrm{out}}
\;+\;
\alpha_{\mathrm{MMD}}\,\mathcal{L}_{\mathrm{MMD}}\Big]
\;+\;
\alpha\!\left(\gamma \lVert \bm \theta \rVert_1
\;+\;
\tfrac{1 - \gamma}{2} \lVert \bm \theta \rVert_2^{2}\right),
\end{aligned}
\label{eq:ctx_loss}
\end{equation}
where $\gamma \in [0, 1]$ balances $L_1$ and $L_2$ penalties, $\alpha \in [0, 1]$ trades off predictive loss against regularization, and $\alpha_{\mathrm{MMD}}\in[0,0.99]$ controls the strength of the balancing loss. All weights ($\alpha, \alpha_{\mathrm{MMD}}, \gamma, \lambda_1, \lambda_2$) were tuned via hyperparameter search (see \Cref{tab:hyperparameters}). During optimization, $\mathcal{L}_{\mathrm{out}}$ and $\mathcal{L}_{\mathrm{MMD}}$ are approximated on mini-batches of patients (\textbf{\cref{sec:methods:mscv}}).

To prevent the balancing penalty $\mathcal{L}_{\mathrm{MMD}}$ from suppressing factual fit early in training, $\alpha_{\mathrm{MMD}}$ was annealed across training: held at zero over the first 20\% of epochs, then linearly ramped to its tunable target value over the next 60\% of epochs, and held constant at that value for the remaining 20\%.

\subsubsection{Multi-source cross-validation and model training} \label{sec:methods:mscv}
CTX was trained on the development set under a three-fold leave-one-cohort-out multi-source cross-validation (MSCV) scheme~\citep{geras_pmlr_2013}. 
In each fold, eleven cohorts (Lodz, METABRIC, Korean Catholic, Gundersen, BCN, Malaysian, Ontario, Omica, TCGA-BRCA, and two of NYU, Cardiff, and Sydney) were used for training, and one of the three cohorts (NYU, Cardiff, or Sydney in turn) was used as the validation set for hyperparameter selection, early stopping, and post-hoc calibration.

When a cohort was used for validation, we further restricted to HR+/HER2$-$ patients who received ET or CET.
We also filtered out patients who received targeted therapies if the information on targeted therapies was available. 
This was done so that model selection and calibration remain consistent with the evaluation cohorts, targeting the subpopulation of interest.
The five evaluation cohorts were strictly held out from all training and model-selection steps. 

\paragraph{Optimization and hyperparameter search.}
All models were trained using the Adam optimizer for up to 100 epochs. 
Each training batch was formed by sampling patients uniformly at random from the pooled development set, with all cohorts concatenated before sampling. 
Sampling was without replacement, with a batch size of 1,100, so each batch comprised 1,100 distinct patients. 
Early stopping was triggered if the validation metric --- the concordance index of the factual survival predictions with respect to observed recurrence outcomes --- did not improve for 15 consecutive epochs.

Within each fold, 100 hyperparameter configurations were evaluated, comprising 50 randomly sampled configurations followed by 50 Bayesian-optimization trials (BoTorch~\citep{balandat2020botorch} Gaussian process with expected improvement acquisition function, exploration parameter $\xi = 1.0$). 
The full hyperparameter search space, covering optimization, modality-specific architecture, and the balancing penalty, is shown in~\Cref{tab:hyperparameters}. 
Training was performed on a single node with eight NVIDIA H200 GPUs for approximately seven hours of wall-clock time for each cross-validation fold. 

\paragraph{Ensembling and calibration.}
After the models were trained, we ensembled the best-performing models and calibrated the ensembled predictions to the five-year horizon. 
Specifically, within each validation fold, the ten hyperparameter configurations with the best validation performance were ensembled by averaging their predictions.
Calibration was performed to ensure that the ensemble predictions were accurate probability estimates, in addition to correctly ranking patients according to their relative risk of recurrence. 
We fit beta calibration to factual ensemble predictions within each validation fold \cite{kull_betacal}, using the five-year recurrence indicator as the target. 
For uncensored patients, this target label was determined by observed five-year recurrence status, whereas for patients censored before five years, event times were imputed by sampling from the population conditional Kaplan--Meier estimator \cite{qi2024conformalized}.
The fitted per-fold calibrators were then applied to the corresponding per-fold predictions on the evaluation cohorts for each treatment arm. 
Finally, the calibrated predictions were averaged across folds to yield a single set of five-year recurrence probability predictions per treatment arm.

\subsubsection{AI genomic risk as an auxiliary clinical input}\label{sec:research_odx}

One of the covariates used as input to the clinical branch of the CTX model is an AI genomic risk score, which is predicted by an auxiliary model. 
Here we describe the training data and procedure used to develop this auxiliary model from the slide and routine clinical variables.

Following prior research implementations~\citep{howard2023integration, li2016mr}, the AI genomic risk score was computed from TCGA bulk RNA-seq data (\texttt{fpkm\_uq\_unstranded}) to infer the expression of a 21-gene assay~\citep{Paik_NEJM_2004}. For each tumor sample, the expression value for each gene was normalized by the upper-quartile expression value, log\textsubscript{2}-transformed, and converted to a z-score.

Training data were accessed from the TCGA GDC Data Portal for the following cancer types: TCGA-LGG, TCGA-LIHC, TCGA-UCEC, TCGA-LUAD, TCGA-BLCA, TCGA-GBM, TCGA-KICH, TCGA-COAD, TCGA-STAD, TCGA-HNSC, TCGA-KIRP, TCGA-LUSC, TCGA-SARC, TCGA-OV, TCGA-KIRC, TCGA-SKCM, TCGA-PRAD, and TCGA-BRCA.
Patients were included if they had at least one WSI and at least one primary-tumor bulk RNA-seq sample available through the GDC Data Portal, from which the target could be computed.
Clinical covariates were defined only for TCGA-BRCA and treated as missing for the other 17 cohorts, because most of them (ER, PR, and HER2 status, IDC/ILC histology, and T/N stage) are breast-specific or carry disease-specific definitions.

The auxiliary model was trained in two stages:
\begin{enumerate}[label=(\roman*)]
    \item Pan-cancer pretraining, using a set comprising 18 of TCGA cohorts (8,309 slides from 6,870 patients), with a batch size of 4,600.
    \item Breast cancer fine-tuning, using the TCGA-BRCA subset (1,108 slides from 1,038 patients), with a batch size of 1,108.
\end{enumerate}

The auxiliary model uses the same architecture and training procedure as CTX (\textbf{\cref{sec:methods:multimodal_architecture}}), with the following modifications:
\begin{enumerate}[label=(\roman*)]
    \item the output head maps to a single output unit (in place of the treatment-specific discrete-time hazard heads)
    \item the training loss was a listwise ranking loss (ListNet~\citep{cao2007learning}), with regularization hyperparameter $\alpha=0$
    \item the target was genomic risk --- specifically a 21-gene recurrence-score formulation~\citep{Paik_NEJM_2004}.
    \item Hyperparameters were selected with the same Bayesian optimization search procedure as for CTX, except that the exploration parameter $\xi$ was set to 0 and the validation criterion was Kendall rank correlation between predicted scores and recorded clinical recurrence scores on a held-out NYU cohort (472 slides from 301 patients with a recorded score).
\end{enumerate}

\subsection{Model evaluation} \label{sec:methods:evaluation}

This section describes how CTX is evaluated. \textbf{\Cref{sec:methods:eval:assumptions}} states the assumptions under which our evaluations carry a causal interpretation, and \textbf{\cref{sec:methods:eval:propensities}} describes the inverse propensity weighting used to adjust for non-random treatment assignment. 
\Cref{sec:methods:eval:external} describes the primary prognostic and predictive evaluations. 
The remaining sections describe secondary analyses: therapeutic decision support (\textbf{\cref{sec:methods:eval:decision}}), explainability (\textbf{\cref{sec:methods:eval:explainability}}), molecular correlates (\textbf{\cref{sec:methods:eval:correlates}}), and pan-cancer transfer (\textbf{\cref{sec:methods:eval:pancancer}}).

\subsubsection{Causal assumptions} \label{sec:methods:eval:assumptions}

Treatment assignment in our observational evaluation datasets was not randomized --- under the standard of care at the time of the treatment decision, patients with certain characteristics are more likely to receive CET than other patients. 
Model evaluations that compare survival outcomes between patients treated with CET and ET are therefore susceptible to confounding.

The predictive evaluations in this appendix nonetheless interpret $\tau(\bm x)$ (\textbf{\cref{sec:methods:notation}}) causally, as the difference in five-year recurrence-free probability that would result from assigning chemoendocrine rather than endocrine-only therapy to a patient with features $\bm x$. 
Our evaluations rest on five conditions on the evaluation data: 
\begin{enumerate}[label=(\roman*)]
    \item Consistency --- each patient's observed outcome equals their outcome under the treatment actually received.
    \item Ignorability --- treatment assignment is independent of the potential outcomes given the measured covariates used by the propensity model (\textbf{\cref{sec:methods:eval:propensities}}).
    \item Treatment positivity --- each patient has a nonzero probability of receiving either treatment given those covariates.
    \item Non-informative censoring --- censoring is independent of the event time given the variables conditioned on in each analysis (treatment arm, CTX-$\tau$ group, and cohort or cancer type where stratified).
    \item Censoring positivity --- within each analysis stratum, patients have a nonzero probability of remaining uncensored through the evaluation horizon.
\end{enumerate}

Conditions (i)--(iii) are the standard identification conditions for treatment-effect estimation from observational data~\citep{rosenbaum1983central, hernan2020whatif}: under them, analyses adjusted by inverse propensity weights (IPW; \textbf{\cref{sec:methods:eval:propensities}}) estimate the corresponding causal quantities. 
Conditions (iv)--(v) are the standard conditions under which the Kaplan--Meier and related estimators are consistent~\citep{kalbfleisch2002statistical}.

\subsubsection{Propensity score estimation and inverse propensity weighting} \label{sec:methods:eval:propensities}

To adjust for non-random treatment assignment (\textbf{\cref{sec:methods:eval:assumptions}}), we weighted each patient in the evaluation set by an estimated inverse propensity weight. 
This assigns greater weight to patient-treatment combinations observed less frequently under the observational assignment mechanism, improving covariate balance between treatment groups~\citep{cole2004adjusted}.

\paragraph{Model.}The propensity score $\pi(\bm x) \coloneqq \mathbb{P}(A=1 \mid \bm X = \bm x) $ is a patient's probability of receiving CET instead than ET, given their clinical covariates. We constructed a propensity score estimator $\hat\pi(\bm x)$ using a Catboost classifier and post-hoc beta calibration \citep{prokhorenkova2018catboost, kull_betacal}. The following clinical covariates were used as features: age at diagnosis (years), menopausal status, MammaPrint score, MammaPrint group, Oncotype DX score, tumor size, number of positive nodes, tumor stage, nodal stage, tumor grade, progesterone receptor status, invasive ductal carcinoma status, invasive lobular carcinoma status, lymphovascular invasion status, race, and cohort ID. These covariates represent established clinical factors likely to influence both treatment assignment and recurrence outcomes in our setting.

\paragraph{Training.} Following the nuisance parameter estimation procedure described by \citet{kennedy2024semiparametric}, we trained the CatBoost portion of $\hat\pi$ using a five-fold cross-fitting scheme. That is, we pooled the five evaluation cohorts (n = 1,994) into a single dataset $\mathcal{D}_{\textrm{ipw}}$ and split it into five folds, stratifying on cohort and the adjuvant chemotherapy label.
This yielded five folds $\mathcal{D}_{\textrm{ipw}}^{(1)}, \dots, \mathcal{D}_{\textrm{ipw}}^{(5)}$, each containing data from all cohorts and both treatment groups. Then, for each $k=1,\dots,5$, we held out one fold $\mathcal{D}_{\textrm{ipw}}^{(k)}$ for inference, and we used the remaining four folds $\{\mathcal{D}_{\textrm{ipw}}^{(j)}\}_{j\neq k}$ for training. Since all folds were used for inference at some point, this process produced inference predictions for all of $\mathcal{D}_{\textrm{ipw}}$, upon which we applied beta calibration. We used default CatBoost hyperparameters to avoid further nested data splitting, as CatBoost achieves strong performance on standard classification tasks without hyperparameter tuning~\citep{catboost_hps1, catboost_hps2}.

\paragraph{Inverse propensity weights.} Letting $\hat\pi_i$ denote our calibrated estimate of $\mathbb{P}(A=1 \mid \bm X =\bm x_i)$, we define the inverse propensity weight (IPW) assigned to patient $i$ as $w_i=1/\hat \pi_i$ if the patient received chemoendocrine therapy ($a_i=1$) and $w_i=1/(1-\hat \pi_i)$ if the patient received only endocrine therapy ($a_i=0$). To reduce variance in downstream IPW-adjusted estimators, we clipped all IPW values greater than 10 to be equal to 10.

\paragraph{Evaluation.} The propensity model predicted the assignment of treatment with an AUC of 0.925. To measure the calibration of our predictions, we grouped our estimated propensity scores into equally sized bins and compared the estimated rates of treatment to observed rates of treatment within each bin. Across 10 quantile bins, the expected calibration error of our propensity estimates was 0.008 and the maximum calibration error was 0.015.

To quantify the effectiveness of our IPW-adjustment in mitigating confounding, we measured the absolute standardized mean difference (SMD) between patients receiving CET and ET for each clinical feature, before and after applying IPW. For categorical features, we treated missingness as a category and we computed the weighted average of one-vs-rest binary SMDs, where weights correspond to the prevalence of each category. For continuous features, we computed a separate binary SMD for missingness, and we reported the weighted average of the continuous SMD and the missingness SMD, where weights are determined by prevalence of missingness. The SMDs for all clinical features used in the propensity model are shown in Extended Data \ref{fig:propensity_scores}. The mean SMD across clinical features decreased from $0.376$ to $0.125$ and the maximum SMD decreased from 1.036 to 0.281 upon IPW adjustment, indicating that IPW reduced the confounding bias in measured covariates.

\subsubsection{Evaluation of CTX as a predictor of prognosis and chemosensitivity} \label{sec:methods:eval:external}

Evaluation of CTX was performed in the held-out evaluation set (n = 1,994 patients with HR+/HER2$-$ early breast cancer from five cohorts).

\paragraph{Prognostic performance.}
CTX-prognostic (\textbf{\cref{sec:methods:notation}}) was evaluated for discrimination, calibration, and association with RFI (without IPW) (see \textbf{\cref{sec:results:prognostic}}). 
Discrimination was assessed by the time-dependent AUC at the 5- and 10-year horizons, computed with the \texttt{tdROC} package~\citep{tdroc}; we report the empirical AUC with 95\% confidence intervals from 1,000 bootstrap replicates, in the pooled evaluation set and within each cohort. 
Calibration was assessed at the same horizons by grouping patients into quintiles of predicted risk; within each quintile, the mean predicted event rate was compared against Kaplan--Meier-estimated observed event rates at the horizon, and the calibration slope and intercept were estimated by ordinary least-squares regression of observed on predicted mean risk across the five bins. 
The association between CTX-prognostic and RFI was estimated by Cox proportional hazards regression on log-transformed CTX-prognostic, with follow-up administratively censored at 5 and 10 years and hazard ratios reported with 95\% confidence intervals. Multivariable analyses additionally adjusted for age at diagnosis, T- and N-stage.

\paragraph{Predictive performance.}
Patients were classified as CTX-$\tau$ high-benefit ($\hat \tau \geq$ two percentage points) or CTX-$\tau$ low-benefit ($\hat \tau <$ two percentage points) (see \textbf{\cref{sec:results:predictive}}).
Within each CTX-$\tau$ category, high-benefit and low-benefit, IPW-adjusted Kaplan--Meier curves displayed RFI outcomes for patients who received CET and those who received ET. 

The principal predictive analysis was a Cox model including dichotomized CTX-$\tau$, an indicator of adjuvant chemotherapy receipt, and their interaction term, weighted by inverse propensity weights $w_i$ (\textbf{\cref{sec:methods:eval:propensities}}), with variance estimated by a robust (sandwich) estimator. The interaction hazard ratio and its p-value quantify whether the association between chemotherapy and RFI varies as a function of the dichotomized CTX-$\tau$ score. The interaction test with dichotomized CTX-$\tau$ was additionally repeated in an unweighted multivariable model adjusted for age at diagnosis, T-stage, N-stage, tumor grade, and histological type (invasive ductal and invasive lobular carcinoma).

An analysis of the relationship between continuous CTX-$\tau$ score and chemotherapy benefit was performed by calculating IPW-adjusted Kaplan--Meier estimators within each CTX-$\tau$ stratum for patients who received CET and those who received ET. 
The within-stratum between-arm contrast was tested by the score test from an IPW-adjusted Cox regression on the treatment indicator, a weighted analog of the log-rank test.

To enable benchmarking, the interaction test was repeated with CTX-$\tau$ as a continuous score, standardized to unit variance with IPW-adjustment, so that hazard ratios were per standard deviation.
The same interaction framework was applied to age, N-stage, T-stage, invasive lobular carcinoma, invasive ductal carcinoma, grade, Ki67 staining score, and Oncotype DX recurrence score.
In addition, the framework was applied for CTX-$\tau$ and the continuous Oncotype DX score in the subset of patients with a recorded Oncotype DX score (n = 983), enabling direct comparison of predictive performance between CTX-$\tau$ and an established genomic assay.

\subsubsection{Therapeutic decision support analysis} \label{sec:methods:eval:decision}

This analysis quantified the recurrence-free rate attainable when chemotherapy is allocated according to CTX-$\tau$ score (see \textbf{\cref{sec:results:decision}}). To do this, we ranked patients by CTX-$\tau$ and estimated the recurrence-free rate achieved by assigning CET to the highest-benefit proportion of patients. Specifically, for $q \in \{0, 10, \dots, 100\}$, the rule assigns CET to q\% of patients in the evaluation set with the highest CTX-$\tau$ scores, and ET to the remainder.

For each treatment-allocation rule, we estimated the five-year recurrence-free rate that would result under that rule. This was computed with an IPW-adjusted Kaplan--Meier estimator among the patients whose observed treatment agreed with the rule's assigned treatment. The estimator used IPW-adjustment (\textbf{\cref{sec:methods:eval:propensities}}) to account for non-random assignment of treatments (\textbf{\cref{sec:methods:eval:assumptions}}). Finally, the recurrence-free rate under observed practice was estimated by the unweighted Kaplan--Meier estimator over the full evaluation set.

The point estimates at each q\% were connected linearly to form a curve, displayed up to a treated proportion of 60\%, which lies within one standard deviation of the mean observed treatment rate across cohorts. Two summary statistics were derived from the curve by linear interpolation between adjacent grid points: the smallest treated proportion matching the recurrence-free rate observed in practice, reported as an absolute and relative reduction in chemotherapy use, and the recurrence rate attained at the observed treated proportion, reported as an absolute and relative reduction in recurrence.

\subsubsection{Pathology foundation model explainability} \label{sec:methods:eval:explainability}

CTX derives much of its input from morphological features extracted by the Falcon foundation model, which are not directly human-interpretable. 
The goal of this analysis was therefore to characterize which morphological patterns in the tissue drive CTX-$\tau$ predictions --- specifically, whether tumors predicted to derive high versus low chemotherapy benefit exhibit distinct, pathologist-recognizable histological features (see \textbf{\cref{sec:results:mechanistic}}).

We conducted this analysis within a single representative evaluation cohort (UNIRAD, n = 362).
First, we analyzed the patch-level embeddings generated by the Falcon foundation model. 
For each $224 \times 224$ patch at $20\times$ magnification of the whole-slide image, Falcon produces a 1,536-dimensional embedding vector. 
For visualization, these high-dimensional embeddings were projected to two dimensions using UMAP~\citep{mcinnes_arxiv_2018}.
To identify recurrent morphological patterns, we performed $k$-means clustering ($k$ = 9) on the 1,536-dimensional patch embeddings; clustering was performed in the embedding space, not on the UMAP coordinates.
For each cluster, we fitted a univariable linear models, regressing the CTX-$\tau$ score on the patient-level cluster proportion. Regression coefficients and p-values were reported for each cluster--endpoint pair, identifying clusters whose morphological content is enriched in tumors predicted to derive high or low chemotherapy benefit.
To enable qualitative assessment of the morphological features captured by each cluster, we sampled representative patches per cluster and arranged them in a grid ordered by the cluster-level CTX-$\tau$-coefficient. Three board-certified pathologists with expertise in breast cancer, blinded to the CTX predictions and clinical outcomes, independently reviewed the UMAP embedding and the representative patch grids, annotating the dominant histomorphological features characterizing the clusters.

\subsubsection{Genomic and transcriptomic correlates} \label{sec:methods:eval:correlates}

To understand the molecular underpinnings of predicted chemotherapy benefit (\textbf{\Cref{fig:explainability}}), we related CTX-$\tau$ to genomic data for TCGA-BRCA samples from cBioPortal~\citep{cerami2012cbio}, using the harmonized GDC data release (see \textbf{\cref{sec:results:mechanistic}}). We restricted all analyses to samples with HR+/HER2$-$ disease (n = 348) to match the clinical setting in which CTX operates. These analyses are exploratory, as TCGA-BRCA contributed to model development.

To characterize the transcriptional programs distinguishing CTX-$\tau$ high- and low-benefit categories, we performed differential gene expression using DESeq2~\citep{love2014moderated}, which models raw count data with a negative binomial generalized linear model. We then ranked genes by their DESeq2 Wald statistic and used this ranking for gene set enrichment analysis, performed with the multilevel algorithm of the \texttt{fgsea} R package~\citep{korotkevich2016fast}, restricting to gene sets with 15--500 members. We tested for enrichment against the Hallmark gene set collection~\citep{liberzon2015molecular}, representing 50 well-defined biological states and processes. We considered Hallmark pathways with a Benjamini--Hochberg adjusted p-value < 0.05 to be significantly enriched. We also specifically tested for enrichment of a cGAS-STING signature~\citep{ashburner2000gene}, in an unadjusted analysis. To relate predicted chemotherapy benefit with genome-wide measures of instability, we computed the Spearman rank correlation of the CTX-$\tau$ score with homologous recombination deficiency (HRD) scores from~\citet{knijnenburg2018genomic}.

\subsubsection{Pan-cancer generalizability analysis} \label{sec:methods:eval:pancancer}

To assess the generalizability of CTX across cancer types (\textbf{\Cref{fig:pancancer}}), we applied the CTX model to cancer types from TCGA: n = 6,692 patients across 16 cancer types for prognostic analysis; n = 5,783 patients across 14 cancer types for predictive analysis (see \textbf{\cref{sec:results:pancancer}}). For each patient, CTX generated both a baseline recurrence risk prediction (CTX-risk) and a chemotherapy-benefit prediction (CTX-$\tau$) as defined in \textbf{\cref{sec:methods:notation}}. Harmonized clinical endpoint information for overall survival (OS) was obtained from the TCGA Pan-Cancer Clinical Data Resource~\citep{liu2018integrated}. Treatment information, including chemotherapy receipt, was also extracted from the GDC clinical data files. Each patient was assigned a binary chemotherapy indicator based on whether any treatment record listed chemotherapy as the treatment type.

\paragraph{Propensity score estimation.} Similarly to \textbf{\cref{sec:methods:eval:propensities}}, we used IPW adjustment to account for non-random treatment allocation in the TCGA observational cohorts. Unlike \textbf{\cref{sec:methods:eval:propensities}}, we trained a separate propensity model for each dataset since we expect treatment assignment mechanisms to vary substantially across cancer types. This required training on small datasets ranging from 269 to 513 patients. In light of small sample sizes, we estimated the propensity scores using logistic regression trained on clinical covariates: age, sex, histological type, histological grade, pathological stage, and clinical stage. Cross fitting was not necessary since logistic regression satisfies the Donsker conditions as described by \citet{kennedy2024semiparametric}. We applied post-hoc beta calibration as described in \textbf{\cref{sec:methods:eval:propensities}}. The resulting inverse propensity weights were applied in the predictive evaluation described below.

For each cancer type, the prognostic association between CTX-risk and OS was assessed using a univariable Cox proportional hazards regression, with CTX-risk as the sole covariate and administrative censoring at 5 years. 
Significance was assessed by a likelihood-ratio test against the covariate-free null; a hazard ratio $> 1$ indicates that higher predicted risk corresponds to worse outcomes. 

The predictive association between CTX-$\tau$ and chemotherapy benefit was evaluated by comparing IPW-adjusted restricted mean survival time (RMST) differences across CTX-$\tau$ groups. Within each cancer type, CTX-$\tau$ was dichotomized into high- and low-benefit groups at the 90th percentile, so that the proportion assigned to the high-benefit group matched that in the breast evaluation set. Within each group, the RMST difference between CET and ET patients was computed as the area between IPW-adjusted Kaplan--Meier survival curves of the two arms, up to the median follow-up time of the cancer type as estimated by the reverse Kaplan--Meier method. The RMST difference quantifies the between-arm contrast within each CTX-$\tau$ group, with larger (positive) values indicating greater chemotherapy benefit. Finally, we computed the difference between the high-benefit and low-benefit RMST-differences, with positive values indicating a greater between-arm contrast in the high-benefit group. 

\end{document}